%% file: main.tex
\documentclass{article}

\usepackage[main, final]{neurips_2026}

\usepackage[utf8]{inputenc} 
\usepackage[T1]{fontenc}    
\usepackage{hyperref}       
\usepackage{url}            
\usepackage{booktabs}       
\usepackage{amsfonts}       
\usepackage{nicefrac}       
\usepackage{microtype}      
\usepackage{xcolor}         
\usepackage{algorithm}
\usepackage{algorithmicx}
\usepackage{algpseudocode}
\usepackage{amsmath}
\usepackage{multirow}
\usepackage{graphicx}
\usepackage{array}
\usepackage{enumitem}

\usepackage{float}
\usepackage{footmisc}

\definecolor{myBlue}{HTML}{0F1A5F}
\definecolor{myRed}{HTML}{721010}
\definecolor{linkpink}{RGB}{220, 50, 130}
\hypersetup{
  colorlinks,
  linkcolor={myRed},
  citecolor={teal},
  urlcolor={linkpink}
}

\title{SEGA: Spectral-Energy Guided Attention for Resolution Extrapolation in Diffusion Transformers}

\renewcommand*{\thefootnote}{\fnsymbol{footnote}}
\vspace{-10pt}
\author{
  Javad Rajabi$^{1,2}$\quad
  Kimia Shaban$^{1,2}$\quad
  Koorosh Roohi$^{1,2,3}$\quad
  David B. Lindell$^{1,2}$\quad
  Babak Taati$^{1,2,3}$\\[2mm]
  $^1$University of Toronto \quad
  $^2$Vector Institute \quad
  $^3$KITE Research Institute\\[2mm]
  \small \texttt{\{rajabi, lindell, taati\}@cs.toronto.edu}\\[2mm]
  \small Project page: \url{https://rajabi2001.github.io/sega/}
  \vspace{-7pt}
}

\begin{document}

\maketitle

\renewcommand*{\thefootnote}{\arabic{footnote}}
\setcounter{footnote}{0}

\input{section/abstract}

\input{section/introduction}


\input{section/related_work}

\input{section/preliminaries}

\input{section/method}

\input{section/analysis}


\input{section/experiments}


\input{section/conclusion}


\section*{Acknowledgements and Disclosures}
\paragraph{Funding and Support.}
This work was conducted as academic research at the University of Toronto
and the Vector Institute and did not receive project-specific funding.
Computational resources for this work were provided through the Vector
Institute, SciNet, and the Digital Research Alliance of Canada.

\paragraph{Competing Interests.}
Javad Rajabi is a research intern at Samsung AI Center Toronto. This work
was conducted independently of that position and received no funding or
resources from Samsung. The authors declare no other competing interests.


\medskip

{
    \small
    \bibliographystyle{unsrt}
    \bibliography{main}
}

\newpage
\input{section/appendix}



\end{document}

%% file: section/abstract.tex
\vspace{-17pt}

\begin{figure}[h]
  \centering
 \includegraphics[width=\textwidth]{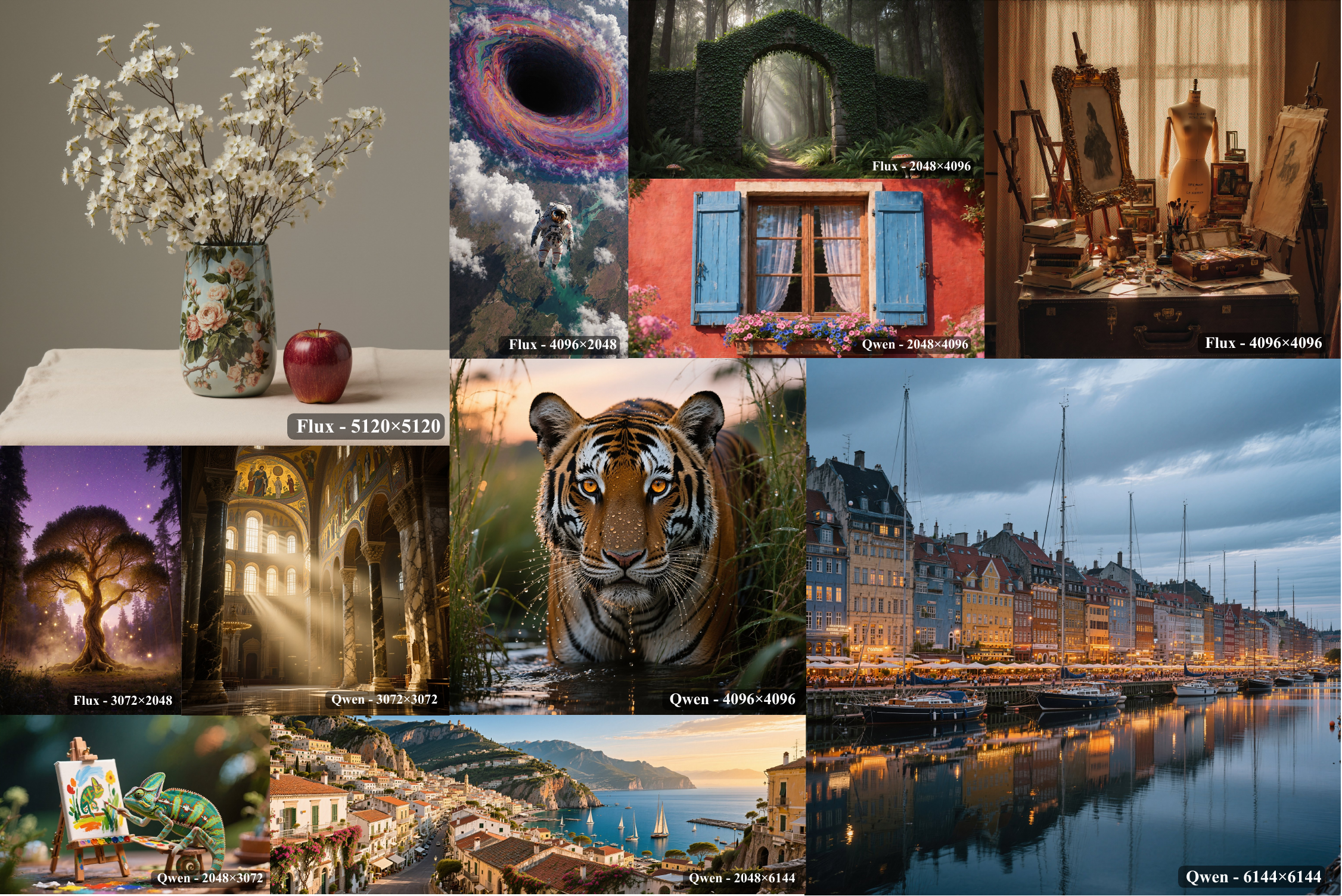}
  \vspace{-15pt}
  \caption{\textbf{Gallery of SEGA.} SEGA unlocks the high-resolution generation capabilities of pre-trained T2I models (Flux~\cite{flux} and Qwen~\cite{qwen}), producing high-quality images. Best viewed zoomed in.}
  \label{fig:teaser}
\end{figure}

\vspace{3pt}
\begin{abstract}
\vspace{-4pt}
    Diffusion transformers (DiTs) have emerged as a dominant architecture for text-to-image generation, yet their performance drops when generating at resolutions beyond their training range. Existing training-free approaches mitigate this by modifying inference-time attention behavior, often through Rotary Position Embeddings (RoPE) extrapolation combined with attention scaling. However, these strategies apply a uniform and content-agnostic scaling across RoPE components with distinct frequency characteristics, inducing a trade-off between preserving global structure and recovering fine detail. We introduce \textbf{SEGA}, a training-free method that dynamically scales attention across RoPE components according to the latent’s spatial-frequency structure at each denoising step. This adaptive scaling improves both structural coherence and fine-detail fidelity. Experiments show that SEGA consistently improves high-resolution synthesis across multiple target resolutions, outperforming state-of-the-art training-free baselines.
\end{abstract}

%% file: section/introduction.tex
\vspace{-5pt}
\section{Introduction}
\label{sec:introduction}
\vspace{-4pt}

Diffusion transformers (DiTs)~\cite{peebles2023scalable, bao2023all} have become the dominant approach to text-to-image (T2I) generation, producing images with a level of quality that would have been hard to imagine just a few years ago.
Despite considerable improvements, existing T2I models remain largely constrained by the resolution ranges used during training, typically between $1024^2$ and $2048^2$ resolutions, limiting their practical applicability~\cite{bu2025hiflow, du2024imax, sigillo2025latent}.
Consequently, extrapolating beyond this training resolution at inference time often leads to notable quality degradation and even structural breakdown.
A straightforward solution is to train or fine-tune models at the target resolution~\cite{hoogeboom2023simple, guo2024make}. However, such approaches are practically limited by the scarcity of high-resolution data, the quadratic cost of longer token sequences, and the need for model-specific fine-tuning. These bottlenecks have motivated growing interest in training-free high-resolution synthesis from pre-trained models~\cite{du2024demofusion, he2023scalecrafter, jin2023training, kim2025diffusehigh}.

Existing training-free methods for high-resolution image generation generally fall into two categories:  (i) direct inference~\cite{zhao2025ultraimage, issachar2025dype, lu2024fit, hou2026boosting} and  (ii) multi-stage guidance-based approaches~\cite{qiu2025freescale, zhang2024frecas, zhang2025diffusion, bu2025hiflow, du2024imax}. Direct inference methods attempt to extend pretrained models to higher resolutions by modifying the denoising process or adjusting components such as positional encoding and attention without additional training. In contrast, multi-stage approaches first generate a base-resolution image and then use it to guide high-resolution synthesis. Although often effective, these methods introduce additional complexity and depend heavily on the quality of the low-resolution prediction. More importantly, they fundamentally cast high-resolution generation as a super-resolution problem, relying on external guidance rather than improving the model's intrinsic ability to extrapolate to higher resolutions.

\begin{figure}[t]
  \centering
  \includegraphics[width=\textwidth]{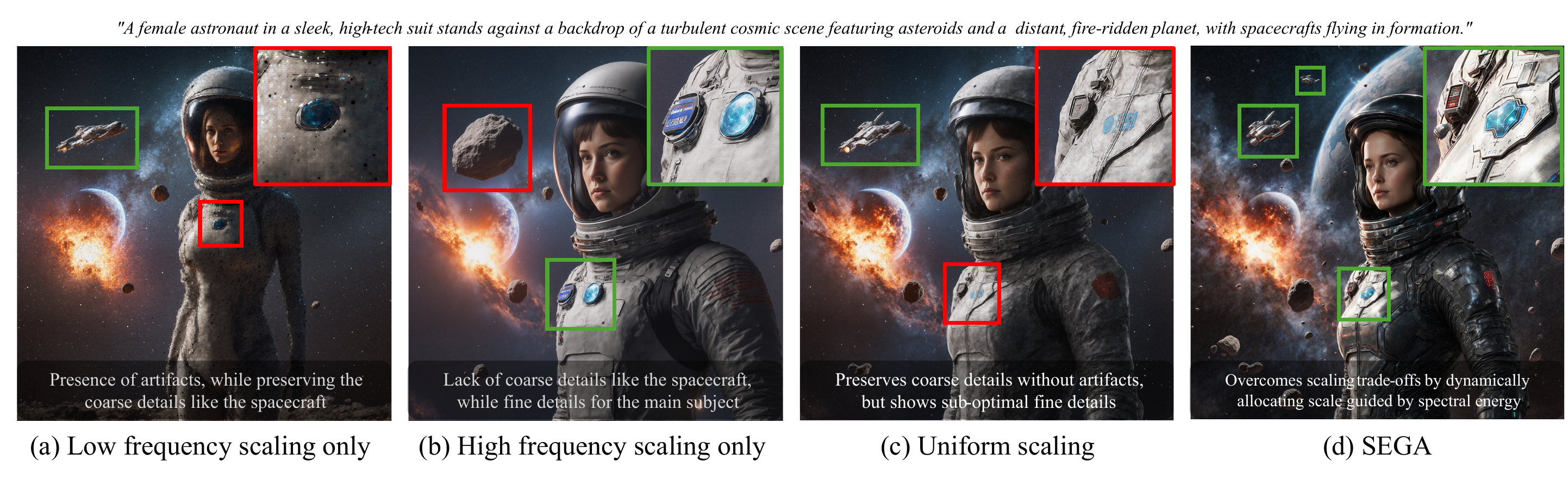}
  \vspace{-17pt}
  \caption{\textbf{Trade-offs in attention scaling at $\mathbf{4096^2}$.}
  RoPE components are coupled to spatial frequencies: low-frequency components support coarse detail and structure, whereas high-frequency components support fine detail and texture. Static scaling fails to balance this trade-off, leading to different failure modes in (a)--(c). SEGA (d) resolves them by dynamically allocating scaling according to spectral energy. {\color[rgb]{0.18,0.55,0.18}Green} and {\color{red}red} boxes indicate successful and failed regions, respectively.
  \vspace{-10pt}}
  \label{fig:scaling_tradeoff}
\end{figure}

In this work, we focus on direct-inference methods for resolution extrapolation in DiTs and address a fundamental failure mode related to positional encoding. When extrapolating pre-trained DiTs to high-resolution synthesis, the relative positional offsets in Rotary Position Embeddings (RoPE)~\cite{rope} deviate significantly from those observed at training time, causing the attention weights to become overly diluted across the expanded token grid. This weakens spatial discrimination in attention and leads to degraded outputs such as blurred textures, repetitive patterns, and structural breakdowns. To counter this, previous approaches, adapted from long-context language modeling, combine RoPE extrapolation with a uniform attention scaling to restore spatial focus~\cite{peng2024yarn}. Specifically, they scale the resulting attention values uniformly across the positional encoding components. While this uniform attention scaling improves image quality, it applies the same adjustment across RoPE components with different frequency characteristics, treating short-wavelength components that govern fine-grained texture identically to long-wavelength components that shape global structure. 
As illustrated in Figure~\ref{fig:scaling_tradeoff}, static scaling induces an inherent trade-off, yielding different failure modes across global structure and fine-grained detail.
The problem is further compounded by two distinct variations in the latent's spectral characteristics. First, the spectral distribution evolves throughout denoising, with the relative contributions of low- and high-frequency bands shifting noticeably as the image resolves from noise to a structured form. Second, the spectral distribution differs across images, depending on their content and structural complexity (e.g., a foggy lake versus a bustling outdoor market). Consequently, a static, uniform scaling at inference time cannot accommodate these variations.

Building on this view, we introduce \textbf{SEGA} (\textbf{S}pectral-\textbf{E}nergy \textbf{G}uided \textbf{A}ttention), a training-free, content-aware method that dynamically adapts attention scaling to the latent's spectral structure by deriving per-component scaling magnitudes at each denoising step.
Our method is motivated by a simple but consequential observation: RoPE components are coupled to spatial frequencies, as shown in Figure~\ref{fig:scaling_tradeoff}. SEGA uses the energy in each corresponding spatial frequency band to determine the scaling applied to each RoPE component: those associated with low-energy bands receive stronger scaling to preserve positional discrimination at those frequencies, whereas components associated with high-energy bands receive weaker scaling to avoid over-amplifying already prominent features. A scalar then controls how strongly this scaling is applied, based on the spectrum's entropy.
The result is an attention scaling that adapts to both the content of the current latent and its evolution across denoising steps, resolving the trade-off induced by fixed global scaling. 

Extensive experiments show that SEGA consistently improves structural coherence and fine-detail fidelity and achieves superior performance across baselines and resolution settings, including ultra-high resolutions exceeding 36 million pixels. SEGA introduces no learnable parameters, requires no fine-tuning or architectural changes, and integrates directly into standard RoPE-based pipelines, making it a minimal yet effective solution for stable high-resolution synthesis across a wide range of extrapolated resolutions, as shown in Figure~\ref{fig:teaser}.

%% file: section/related_work.tex
\vspace{-7pt}
\section{Related Work}

\vspace{-3pt}
\subsection{High-Resolution Image Synthesis}
\vspace{-3pt}
Preserving both global structure and fine-grained detail remains an open challenge in high-resolution generation. Training-based approaches address this through progressive upsampling~\cite{ho2022cascaded, gu2023matryoshka, skorokhodov2024hierarchical, haji2025improving}, latent-space super-resolution~\cite{jeong2025latent}, or explicit retraining on high-resolution data or model-specific fine-tuning like Diffusion-4K~\cite{zhang2025diffusion}. By contrast, training-free methods~\cite{zhang2024frecas, wu2025megafusion, lin2024accdiffusion, huang2024fouriscale} adapt pretrained models at inference time. In U-Net architectures, methods such as DemoFusion~\cite{du2024demofusion}, FreeScale~\cite{qiu2025freescale}, and FreCaS~\cite{zhang2024frecas} improve high-resolution generation through patch stitching, multi-scale fusion, or cascaded sampling, but often introduce additional inference complexity. In DiTs, training-free extrapolation has largely relied on more complex strategies, often involving two-stage pipelines in which a base-resolution trajectory guides high-resolution sampling, as in I-Max~\cite{du2024imax}, HiFlow~\cite{bu2025hiflow}, and ScaleDiff~\cite{koh2025scalediff}. While effective, these methods depend on multi-stage guidance and often introduce additional complexity into the denoising process.

\vspace{-3pt}
\subsection{RoPE-based Length Extrapolation}
\vspace{-3pt}
The challenge of high-resolution generation in DiTs closely mirrors long-context extrapolation in large language models (LLMs)~\cite{ding2024longrope, hu2025pepe}, largely driven by advances in RoPE~\cite{rope}. Standard training-free methods~\cite{chen2023extending, bloc97_ntk, peng2024yarn} formulate extrapolation as recalibration of RoPE's rotary frequencies. Position Interpolation~\cite{chen2023extending} compresses position indices to fit longer sequences within the training range, limiting phase drift. NTK~\cite{bloc97_ntk} adjusts the RoPE base frequency to redistribute positional variation more evenly across dimensions, thereby improving extrapolation to longer sequences. YaRN~\cite{peng2024yarn} builds on both by applying frequency-band-specific interpolation strategies and introducing an additional uniform attention scaling. Recent works adapt these principles to visual domains~\cite{zhao2025ultravico, zhao2025riflex}. DyPE~\cite{issachar2025dype} introduces step-wise, time-aware positional adjustments across the diffusion timesteps. UltraImage~\cite{zhao2025ultraimage} alleviates repetitive artifacts by shifting the dominant frequency to align with the training resolution and employing entropy-guided attention concentration. However, these approaches largely rely on predefined heuristics or target-resolution alignments. In contrast, our method directly analyzes the spectral energy of the intermediate latent to dynamically adjust attention scaling. By amplifying high-energy bands and suppressing low-energy ones, it preserves fine-grained detail without compromising structural fidelity. See Appendix~\ref{app-related-work} for more detailed related work.

%% file: section/preliminaries.tex
\vspace{-5pt}
\section{Preliminaries}
\label{sec:preliminaries}
\vspace{-3pt}

\vspace{-3pt}
\paragraph{Rotary Position Embedding (RoPE)}
Positional embeddings provide spatial priors for transformer architectures, which form the core of DiT models. They encode coordinate information into feature representations, addressing the models' inherent permutation equivariance.
Among various designs, RoPE~\cite{rope} is a widely used scheme that encodes relative positions through rotation in the embedding space, and
it has been adopted in recent T2I models such as Flux~\cite{flux} and Qwen~\cite{qwen}.

RoPE encodes a position $n$ by applying a series of 2D rotations to paired dimensions, each at a distinct angular frequency determined by the embedding dimension index. Given a vector $\mathbf{x} \in \mathbb{R}^D$ at position $n$, RoPE partitions $\mathbf{x}$ into $D/2$ two-dimensional subspaces and rotates the $d$-th subspace as
\begin{equation}
\boldsymbol{f}^{\text{RoPE}}(\mathbf{x}, n, \boldsymbol{d}) = 
\begin{bmatrix}
    \cos(n\theta_d) & -\sin(n\theta_d) \\
    \sin(n\theta_d) & \phantom{-}\cos(n\theta_d)
\end{bmatrix}
\begin{bmatrix}
    x_{2d} \\
    x_{2d+1}
\end{bmatrix},
\end{equation}
where $\boldsymbol{\theta} \in \mathbb{R}^{D/2}$ with $\theta_d = b^{-2d/D}$ for $d = 0, \dots, D/2 \:-\: 1$ and $b = 10{,}000$. In practice, RoPE is applied to the query and key vectors before the dot product operation in the attention mechanism. Additionally, it can be shown that the dot product of two RoPE-embedded vectors depends only on their relative distance, so attention naturally encodes relative positional information. For 2D images, RoPE is typically applied axially: half of the hidden dimensions encode horizontal positions and the other half encode vertical positions, enabling independent offsets along each axis~\cite{heo2024rotary}.

\vspace{-5pt}
\subsection{Length Extrapolation Techniques and Attention Scaling}
\label{sec:temperature}

Although RoPE provides an effective positional bias within the training, models that rely on it often degrade at unseen resolutions, where attention must operate on out-of-distribution positional offsets.
Several methods have been proposed to adapt RoPE to longer sequences at inference time, given an extrapolation ratio $s = (L_{\text{target}} / L_{\text{train}})$, where $s > 1$. \emph{Position Interpolation} (PI)~\cite{chen2023extending} linearly compresses position indices via $n \mapsto n/s$ for position $n$, which uniformly transforms all RoPE components to $\theta_d / s$ so extrapolated positions remain within the training range. \emph{NTK-aware}~\cite{bloc97_ntk} instead adjusts $b$ to $b' = b \cdot s^{D/(D-2)}$, which stretches the angular frequency of each rotary dimension $\theta_d$. \emph{YaRN}~\cite{peng2024yarn} unifies these ideas by partitioning rotary dimensions and applying a gradual interpolation-extrapolation strategy, a.k.a.\ \textit{NTK-by-parts}~\cite{peng2024yarn}. Specifically, it smoothly interpolates the modified frequencies as $\theta_d' = (1-\lambda_d)\frac{\theta_d}{s} + \lambda_d \theta_d$ using a ramp function $\lambda_d \in [0,1]$.

Another key component of YaRN is \textit{attention scaling}, applied to the logits before the softmax. Notably, this effect can be implemented through RoPE by scaling the query and key vectors after rotation, thereby changing the effective attention behavior without altering the attention mechanism itself~\cite{peng2024yarn}. 
YaRN proposes a constant logit scaling factor $\tau(s)$ to compensate for the change in attention behavior under extrapolation, modifying attention as
\begin{equation}
    \mathrm{Attn}(\mathbf{Q}, \mathbf{K}, \mathbf{V})
    = \mathrm{softmax}\!\left(\tau(s)\cdot\frac{\mathbf{Q}\mathbf{K}^\top}{\sqrt{d_k}}\right)\mathbf{V},
    \qquad
    \tau(s) = 0.1\ln(s) + 1
\end{equation}
where $\mathbf{Q}$, $\mathbf{K}$, and $\mathbf{V}$ represent the query, key, and value matrices, respectively; $d_k$ denotes the dimensionality of the queries and keys. The scaling factor $\tau(s)$ was determined empirically for length extrapolation in language models by minimizing perplexity~\cite{peng2024yarn}. The same heuristic has since been adopted in image generation~\cite{lu2024fit}. However, this scaling remains uniform across all RoPE frequencies. Since different RoPE dimensions exhibit distinct characteristics and contribute unevenly to spatial structure, a constant scaling factor is suboptimal; it may over-sharpen some spatial-frequency bands while over-smoothing others, motivating a dynamic scaling strategy. 

%% file: section/method.tex
\vspace{-5pt}
\section{Method}
\label{sec:sega}

\vspace{-5pt}
\textit{Spectral-Energy Guided Attention} (SEGA) introduces content-aware dynamic scaling into DiTs by coupling lightweight spectral analysis with RoPE components.
Our key insight is that RoPE scaling for high-resolution extrapolation should be content-aware rather than fixed and uniform. SEGA achieves this by deriving per-dimension scaling from the latent's spectral structure at each denoising step.

\vspace{-3pt}

\paragraph{Formulation Overview.}
SEGA applies attention scaling through RoPE using a dimension-wise scaling term $m_d$. Specifically, for a token at position $n$ along axis $a$, we define
\begin{equation}
    \boldsymbol{f}^{\text{SEGA}}(\mathbf{x}, n, d)
    = m_d^{(a)} \cdot \boldsymbol{f}^{\text{RoPE}}(\mathbf{x}, n, d),
    \qquad
    m_d^{(a)} = m_{\text{ref}} \cdot \mathcal{M}_d^{(a)}(\mathbf{Z}),
    \label{eq:sega_rope}
\end{equation}
where $m_{\text{ref}}$ is a scalar determined by the target resolution. Here, $\mathcal{M}_d^{(a)}(\mathbf{Z})$ is our novel dynamic modulator derived from the spectral structure of the current intermediate latent $\mathbf{Z}$. It consists of two complementary components: $s_d^{(a)}(\mathbf{Z})$, a \emph{per-dimension correction} that determines the distribution of scaling across RoPE dimensions, and $\sigma(\mathbf{Z})$, a \emph{global amplitude factor} that sets the strength of that adjustment. The remainder of this section describes how spectral structure is extracted from $\mathbf{Z}$ (Section~\ref{sec:analysis}) and converted into $s_d^{(a)}$ and $\sigma$ to assemble the final formula (Section~\ref{sec:allocation}).

\subsection{Spectral Analysis of the Latent}
\label{sec:analysis}

The first stage of SEGA transforms the current latent from the spatial domain to the frequency domain to characterize the spatial frequency content. Given the latent hidden states $\mathbf{Z} \in \mathbb{R}^{N \times C}$ with $N = H \cdot W$ tokens,\footnote{For notational simplicity, we omit the batch dimension $B$ in our formulation, as all operations are applied independently across the batch.} we reshape them back to their native 2D layout, average across channels, and subtract the average value across the spatial dimensions to obtain a zero-centered 2D map $\tilde{\mathbf{M}} \in \mathbb{R}^{H \times W}$ that summarizes the spatial structure of the latent. 
From $\tilde{\mathbf{M}}$ we extract two complementary spectral views from a single 2D Fast Fourier Transform $\mathcal{F}_{2\mathrm{D}}$:
\begin{itemize}
    \item \textbf{Axis-wise profiles.} For each axis $a \in \{H, W\}$ with length $L_a$, we marginalize the 2D power spectrum $\left|\mathcal{F}_{2\mathrm{D}}[\tilde{\mathbf{M}}]\right|^2$ over the orthogonal frequency axis to obtain a 1D profile $\mathcal{E}_a \in \mathbb{R}^{\lfloor L_a / 2 \rfloor}$. Each profile maps spectral energy to spatial frequencies along its axis.
    \item \textbf{Radial profile.} We obtain $\mathcal{E}_{\text{iso}}$ by averaging the same 2D power spectrum within concentric rings. This profile discards directional information and instead provides a rotation-invariant summary of how energy is distributed across spatial scales.
\end{itemize}

These profiles then determine the scaling of each RoPE dimension. Because RoPE is applied separately along the height and width axes, the axis-wise profiles capture directional differences in spectral energy and allow the corresponding RoPE dimensions to be scaled independently, while the radial profile determines the strength of this scaling, as described in the next section.

\vspace{-3pt}
\subsection{From Spectrum to Per-Dimension RoPE Scaling}
\label{sec:allocation}

The second stage converts the spectral profiles into the modulator $\mathcal{M}(\mathbf{Z})$, which defines the per-dimension scaling applied to the rotary embeddings. This formulation consists of three components: a reference scale that anchors the scaling, a per-dimension term that scales individual dimensions, and a global gate that controls the strength of that scaling.

\paragraph{Reference scale.}
The reference scale $m_{\text{ref}}$ is a scalar determined solely by the ratio between the target and training resolutions. Assuming $R_{\text{target}} / R_{\text{train}} \geq 1$, we adopt a power-law form,
\begin{equation}
    m_{\text{ref}} = \left( \frac{R_{\text{target}}}{R_{\text{train}}} \right)^{\kappa},
    \label{eq:m_ref}
\end{equation}
where  $\kappa > 0$ is a small exponent chosen empirically. See Appendix~\ref{app:baseline_ablation} for alternative formulations.

\vspace{-3pt}
\paragraph{Per-dimension correction.}
Each RoPE dimension governs the attention mechanism's sensitivity at a specific spatial wavelength, modifying the scaling at dimension $d$ directly alters how sharply the model can discriminate positional offsets at that wavelength, and therefore affects the corresponding spatial frequency. This coupling motivates a per-dimension correction tied to the latent's actual spectral content. For each RoPE dimension $d$ on axis $a$, we use its wavelength $T_d = 2\pi/\theta_d$ to identify the corresponding band in $\mathcal{E}_a$, retrieve the log-energy $\hat{E}^{(a)}_d$, and standardize it across dimensions as $z^{(a)}_d = (\hat{E}^{(a)}_d - \mu^{(a)}) / \nu^{(a)}$ , where $\mu^{(a)}$ and $\nu^{(a)}$ denote the mean and standard deviation of $\hat{E}^{(a)}$. To enforce a strict zero-sum redistribution, the final correction is defined as $s^{(a)}_d = \phi(z^{(a)}_d) - \mathbb{E}[\phi(z^{(a)})]$, where $\phi(\cdot)$ is a non-linearity, for which we use $\tanh$.
By construction, $s^{(a)}_d < 0$ when dimension $d$ falls in a band with below-average energy and $s^{(a)}_d > 0$ when it falls in a band with above-average energy, while the zero-mean property $\sum_d s^{(a)}_d = 0$ ensures that the correction adjusts the scaling across dimensions without shifting its overall average.

\vspace{-5pt}
\paragraph{Global amplitude factor.} 
To regulate the \emph{magnitude} of the scaling introduced by the axis profiles, SEGA reduces the radial profile $\mathcal{E}_{\text{iso}}$ to a single scalar statistic that captures whether the latent's spectral energy is concentrated in a few dominant bands or spread evenly across all bands. For this purpose we adopt the \emph{spectral flatness}, also known as the \emph{Wiener entropy}, defined as the ratio of the geometric mean to the arithmetic mean of a power spectrum. Applied to $\mathcal{E}_{\text{iso}}$, this yields
\begin{equation}
    \mathrm{SF}\!\left(\mathcal{E}_{\text{iso}}\right) 
     = \frac{\exp\!\left( \frac{1}{n_{\text{bins}}^{(\text{iso})}}\sum_{b=0}^{n_{\text{bins}}^{(\text{iso})} - 1} \ln \mathcal{E}_{\text{iso}}[b] \right)}
               {\frac{1}{n_{\text{bins}}^{(\text{iso})}}\sum_{b=0}^{n_{\text{bins}}^{(\text{iso})} - 1} \mathcal{E}_{\text{iso}}[b]}
    \in (0, 1],
\end{equation}
where $n_{\text{bins}}^{(\text{iso})}$ is the number of radial bins used to compute $\mathcal{E}_{\text{iso}}$. We then remap  the spectral flatness through a simple nonlinearity to produce a scalar \emph{amplitude factor}:
\begin{equation}
    \sigma = 1 - \mathrm{SF}(\mathcal{E}_{\text{iso}})^{\gamma} \in [0, 1],
    \label{eq:sigma}
\end{equation}
where $\gamma \geq 1$ controls how quickly $\sigma$ rises as the spectrum departs from flatness. Without clear spectral structure, $\sigma \to 0$ and SEGA suppresses its scaling; as structural content resolves, $\sigma \to 1$ and the correction applies at full strength.

\paragraph{Final scaling formula.}
Combining the three components, we define the modulator and the resulting per-dimension scaling $m^{(a)}_d$ along each spatial axis $a \in \{H, W\}$ as
\begin{equation}
    \mathcal{M}^{(a)}_d(\mathbf{Z}) = 1 - \sigma \cdot s^{(a)}_d,
    \qquad
    m^{(a)}_d = m_{\text{ref}} \cdot \mathcal{M}^{(a)}_d(\mathbf{Z}).
    \label{eq:sega_mscale}
\end{equation}
 Intuitively, $m_{\text{ref}}$ sets the shared magnitude across RoPE dimensions, $s_d^{(a)}$ determines which dimensions are scaled above or below that reference, and $\sigma$ controls the strength of this redistribution. In this way, SEGA adapts continuously to the latent's spectral content at each denoising step, sharpening attention at under-resolved frequencies and softening it at over-emphasized ones.

%% file: section/analysis.tex
\begin{figure}[t]
  \centering
  \includegraphics[width=\textwidth]{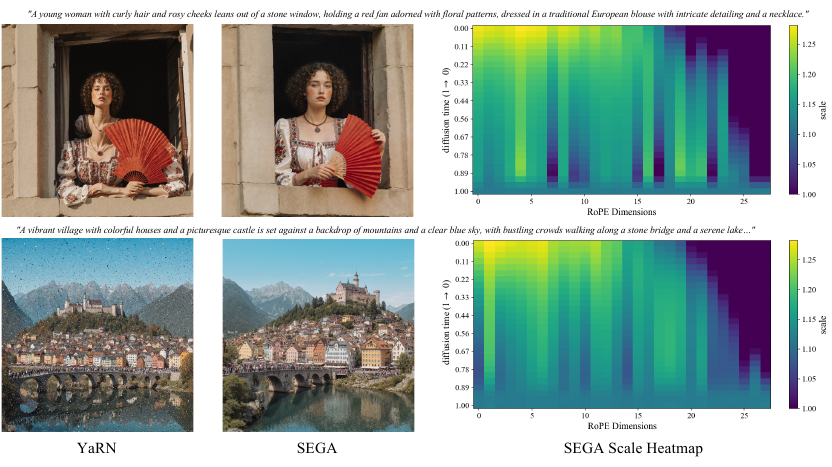}
  \vspace{-17pt}
  \caption{\textbf{SEGA scaling maps at $\mathbf{4096^2}$.} For two representative prompts, the scaling maps show how the horizontal-axis scaling magnitudes $m_d$ change across RoPE dimensions over denoising time.
  \vspace{-15pt}}
  \label{fig:scale_heatmap}
\end{figure}

\vspace{-7pt}
\section{Analysis of Spectral-Energy Guided Attention}
\label{main-analysis}
\vspace{-5pt}

To better understand how SEGA and spectral guidance influence denoising, we analyzed scaling behavior and the attention focus during the denoising process. As shown in Figure~\ref{fig:scale_heatmap}, we visualized the resulting \textit{scaling map}, a temporal representation of how the attention scaling factors $m_d$ are distributed throughout the denoising process. When comparing the scaling maps produced for two distinct prompts, as shown, the difference is apparent. The method yields a customized scaling map for each image, effectively acting as a unique \textit{spectral fingerprint}. This occurs because SEGA is content-aware, dynamically adapting scaling to the latent's spatial frequencies. In early steps where the latent is dominated by noise and the spectrum is relatively flat, the scaling remains near the reference scale $m_{\text{ref}}$. However, as distinct structural energy emerges in later steps, SEGA selectively redistributes scaling across RoPE dimension $d$ to sharpen focus at under-resolved spatial frequency bands while softening it at over-emphasized ones.

\begin{figure}[t]
  \centering
  \includegraphics[width=\textwidth]{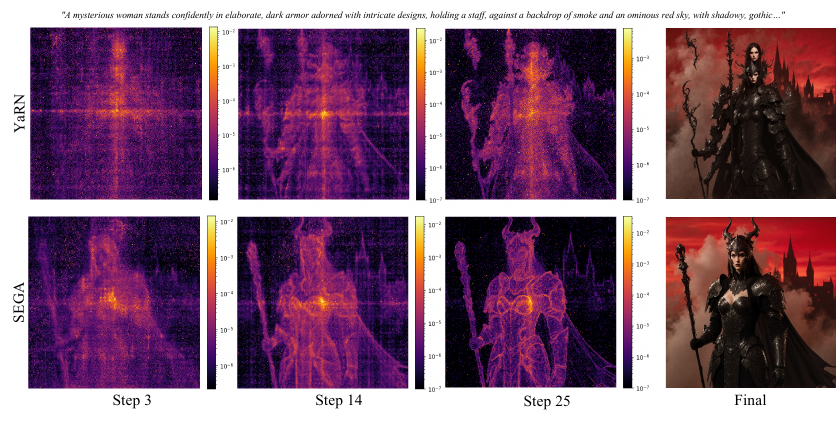}
  \vspace{-19pt}
  \caption{\textbf{Impact on Attention Evolution.} Visual comparison of attention maps for the center latent token in YaRN and SEGA across multiple denoising steps, evaluated on Flux at $4096^2$. 
  \vspace{-10pt}}
  \label{fig:attention_map}
\end{figure}

This content-aware spectral redistribution directly impacts the attention mechanism's stability. As visualized in Figure~\ref{fig:attention_map}, YaRN~\cite{peng2024yarn}, which uses fixed, uniform scaling, suffers from attention dilution, where the model loses the ability to discriminate between positional offsets. SEGA mitigates this failure mode by shaping the attention grid much earlier in the denoising process. By dynamically modulating the magnitude of rotary embeddings, our method preserves semantic locality and entity consistency that uniform scaling methods fail to maintain.

%% file: section/experiments.tex
\vspace{-5pt}
\section{Experiments}
\label{sec:exps}

\input{tables/flux-aesthetic-main}

\begin{figure}[t]
  \centering
  \includegraphics[width=\textwidth]{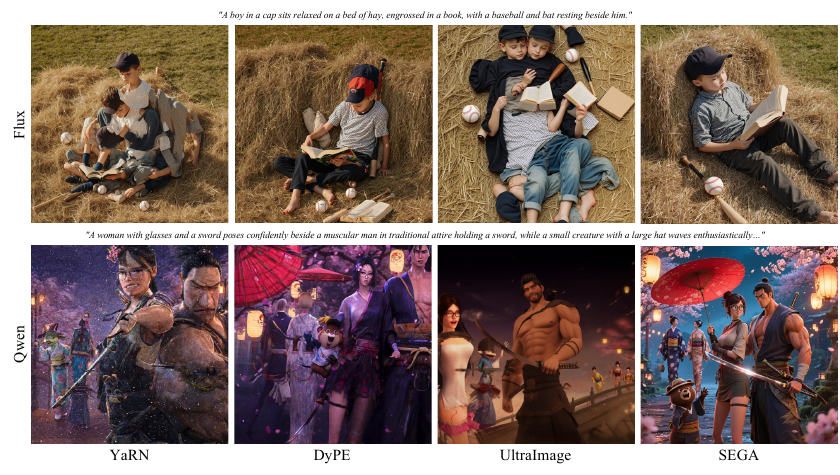}
  \vspace{-15pt}
  \caption{\textbf{Qualitative comparison.} Results on two representative prompts for Qwen and Flux at $4096^2$ resolution show that SEGA improves structural coherence and fine detail over other methods.
  \vspace{-10pt}}
  \label{fig:qualitative}
\end{figure}

\input{tables/qwen-aesthetic}

\vspace{-5pt}
\paragraph{Experimental Settings.} We evaluated our proposed method, \textbf{SEGA} on both Flux~\cite{flux} and Qwen~\cite{qwen}. Throughout the paper, we use NTK~\cite{bloc97_ntk} as the default length extrapolation method for SEGA, unless explicitly stated otherwise. Across all experiments, we set $\gamma$ to 1.5 and $\kappa$ to 0.08.

\vspace{-5pt}
\paragraph{Baselines.} We evaluated SEGA across both the Flux~\cite{flux} and Qwen~\cite{qwen} architectures. We compared our method against two primary categories: direct inference techniques (NTK~\cite{bloc97_ntk}, YaRN~\cite{peng2024yarn}, DyPE~\cite{issachar2025dype}, and UltraImage~\cite{zhao2025ultraimage}), multi-stage guidance approaches (HiFlow~\cite{bu2025hiflow}, I-Max~\cite{du2024imax}, and ScaleDiff~\cite{koh2025scalediff}). Note that the multi-stage guidance methods are exclusively evaluated on Flux to align with their official implementations. See Appendix~\ref{additional-results} for additional methods.

\vspace{-3pt}
\paragraph{Evaluation.} We used prompts and reference images from the Aesthetic-4K~\cite{zhang2025diffusion} dataset. We also curated a ``Zero-Shot'' benchmark comprising detailed prompts generated by an LLM, with results provided in the Table~\ref{tab:zeroshot_resolution}. Quantitative experiments are conducted across four high-resolution configurations: $2048 \times 4096$, $4096 \times 2048$, $3072^2$, and $4096^2$. We evaluate image quality using FID~\cite{fid} and the reference-free metrics MUSIQ (MSQ)~\cite{ke2021musiq}, and CLIP-IQA (CQA)~\cite{wang2023exploring}. Semantic alignment is measured by CLIP Score (CS)~\cite{clipscore,hessel2021clipscore}, while joint alignment and human-preferred visual quality are assessed using ImageReward (IR)~\cite{xu2023imagereward}, PickScore (PS)~\cite{kirstain2023pick}, and HPSv2~\cite{wu2023human}.

\vspace{-5pt}
\subsection{Comparison to State-of-the-Art Methods}

\paragraph{Qualitative comparison.}
When extrapolated to high resolutions, current direct-inference methods (e.g., YaRN~\cite{peng2024yarn}, DyPE~\cite{issachar2025dype}, and UltraImage~\cite{zhao2025ultraimage}) often suffer from severe structural degradation, visual artifacts, and semantic omissions. As shown in Figure~\ref{fig:qualitative}, SEGA better preserves global structural coherence, fine-grained semantic fidelity, and overall visual quality across both the Flux and Qwen architectures, even for complex prompts.

\vspace{-5pt}
\paragraph{Quantitative comparison.} 
As shown in Table~\ref{tab:flux_compact} and Table~\ref{tab:resolution_all_metrics}, SEGA establishes a new state-of-the-art for high-resolution image generation across both the Flux and Qwen architectures. On the Flux model, SEGA consistently achieves the highest semantic alignment and image quality across different settings. The evaluation on the Qwen model further validates these findings. Notably, at the $4096^2$ resolution, SEGA outperforms all baseline models across every evaluated metric, setting a new benchmark for high-resolution generation.

Beyond overall image quality, SEGA exhibits robustness and consistency across a diverse range of high resolutions, including non-square aspect ratios. While other models experience significant performance drops as the resolution increases, SEGA maintains highly stable results. This shows that SEGA extends generation capabilities far beyond the training resolutions of the base models.

\input{tables/flux-aesthetic-ablation-4k}

\vspace{-5pt}
\subsection{Ablation Study}
\vspace{-3pt}
To validate our design choices, we conduct a comprehensive ablation study on the Flux architecture at the $4096^2$ resolution, as detailed in Table~\ref{tab:flux_ablation_4k}. First, we evaluate the core necessity of dynamic spectral guidance by comparing SEGA against the same baseline using NTK~\cite{bloc97_ntk} but with fixed scaling. SEGA yields substantial improvements across all metrics, confirming that fixed scaling fails to maintain structural integrity at extreme resolutions. Next, we ablate the design of our guidance mechanism by restricting SEGA to either \textit{Axis-only} or \textit{ Global-only} scaling. While applying either axis-specific scaling or global scaling independently provides substantial improvements over the baseline, both fall short of the complete method. Finally, we ablate our default choice of NTK~\cite{bloc97_ntk} as the base length extrapolation method by substituting it with YaRN~\cite{peng2024yarn} and DyPE~\cite{issachar2025dype}.

%% file: tables/flux-aesthetic-main.tex
\begin{table}[t]
\centering
\scriptsize
\caption{Comparison of SEGA against state-of-the-art baselines on Flux across four high-resolution settings on Aesthetic-4K~\cite{zhang2025diffusion}. Best and second-best results are shown in \textbf{bold} and \underline{underlined}.}
\label{tab:flux_compact}
\setlength{\tabcolsep}{3pt}
\renewcommand{\arraystretch}{1.0}
\resizebox{\textwidth}{!}{
\begin{tabular}{@{} l | cccccc | cccccc @{}}
\toprule
\multirow{2}{*}{Method} & \multicolumn{6}{c|}{2048 $\times$ 4096} & \multicolumn{6}{c}{4096 $\times$ 2048} \\
\cmidrule(lr){2-7} \cmidrule(l){8-13}
& IR$\uparrow$ & PS$\uparrow$ & CS$\uparrow$ & MSQ$\uparrow$ & FID$\downarrow$ & FID$_p$$\downarrow$ & IR$\uparrow$ & PS$\uparrow$ & CS$\uparrow$ & MSQ$\uparrow$ & FID$\downarrow$ & FID$_p$$\downarrow$ \\
\midrule
Base         & 0.39  & 21.64 & 27.32 & 52.96 & 158.86 & 67.31  & -0.74 & 20.61 & 26.99 & 50.92 & 173.18 & 70.65 \\
\midrule
HiFlow       & 1.14  & 22.61 & 27.72 & 44.30 & \underline{152.24} & 69.23  & \underline{0.89}  & 22.31 & 28.00 & 43.45 & 169.60 & 69.35 \\
I-Max        & 1.11  & 22.71 & 27.60 & 32.58 & 157.82 & 68.21  & 0.87  & \underline{22.54} & 28.30 & 38.71 & 162.30 & 66.92 \\
ScaleDiff    & \underline{1.17}  & 22.84 & \underline{29.03} & \textbf{55.84} & 157.62 & 72.21  & \textbf{0.97}  & 22.51 & 28.58 & \textbf{56.85} & 162.38 & 77.65 \\
\midrule
PI           & -0.83 & 19.77 & 22.87 & 38.73 & 235.83 & 182.50 & -1.10 & 19.50 & 23.05 & 37.48 & 225.65 & 179.96 \\
NTK          & 0.80  & 21.96 & 27.75 & 48.32 & 157.21 & 66.86  & 0.54  & 22.01 & 27.93 & 48.90 & 156.22 & \underline{61.84} \\
YaRN         & 0.97  & 22.63 & 28.48 & 52.30 & 156.97 & 83.35  & 0.30  & 21.82 & 27.99 & 52.10 & \underline{154.44} & 76.42 \\
DyPE         & 1.10  & \underline{22.90} & 28.87 & 53.35 & 159.81 & 85.12  & 0.53  & 22.15 & 28.26 & 52.85 & 158.49 & 85.39 \\
UltraImage   & 0.82  & 22.41 & 28.57 & \underline{55.40} & 157.33 & \textbf{59.19}  & 0.60  & 21.99 & \underline{28.84} & \underline{55.13} & 157.97 & 63.11 \\
\midrule
SEGA         & \textbf{1.21}  & \textbf{22.91} & \textbf{29.18} & 53.65 & \textbf{151.93} & \underline{64.54}  & 0.86  & \textbf{22.58} & \textbf{28.99} & 53.30 & \textbf{153.10} & \textbf{55.85} \\
\bottomrule

\toprule
\multirow{2}{*}{Method} & \multicolumn{6}{c|}{3072 $\times$ 3072} & \multicolumn{6}{c}{4096 $\times$ 4096} \\
\cmidrule(lr){2-7} \cmidrule(l){8-13}
& IR$\uparrow$ & PS$\uparrow$ & CS$\uparrow$ & MSQ$\uparrow$ & FID$\downarrow$ & FID$_p$$\downarrow$ & IR$\uparrow$ & PS$\uparrow$ & CS$\uparrow$ & MSQ$\uparrow$ & FID$\downarrow$ & FID$_p$$\downarrow$ \\
\midrule
Base         & 0.49  & 23.00 & 28.18 & 49.93 & 162.25 & 68.63 & -0.72 & 20.31 & 25.34 & 26.52 & 183.33 & 121.93 \\
\midrule
HiFlow       & 1.26  & \underline{23.22} & 28.48 & 43.45 & 154.32 & 64.72 & \underline{1.26}  & \underline{23.17} & 28.40 & 33.37 & 155.56 & \underline{60.70} \\
I-Max        & 1.28  & 23.15 & 28.42 & 35.23 & \underline{151.98} & 64.56 & \underline{1.26}  & 23.10 & 28.45 & 26.36 & \underline{151.74} & 64.06 \\
ScaleDiff    & \underline{1.30}  & \underline{23.22} & 28.73 & \textbf{53.58} & 153.73 & 72.92 & 1.23  & 23.16 & 28.64 & \underline{44.51} & 153.05 & 76.14 \\
\midrule
PI           & 0.19  & 21.03 & 25.64 & 48.61 & 200.04 & 155.60 & -0.28 & 20.54 & 24.16 & 29.09 & 208.48 & 202.01 \\
NTK          & 0.90  & 22.47 & 28.27 & 35.99 & 156.10 & 63.25 & -0.29 & 20.82 & 25.52 & 23.85 & 182.29 & 138.61 \\
YaRN         & 1.11  & 22.89 & 29.10 & 50.49 & 157.04 & 84.31 & 0.88  & 22.21 & 28.30 & 42.87 & 160.48 & 98.52 \\
DyPE         & 1.21  & 23.15 & 29.17 & 51.71 & 156.91 & 82.45 & 1.01  & 22.56 & \underline{28.79} & 43.23 & 156.21 & 97.88 \\
UltraImage   & 1.17  & 22.65 & \underline{29.30} & \underline{53.56} & 155.92 & \underline{61.61} & 0.61  & 21.74 & 28.16 & 43.61 & 167.04 & 63.91 \\
\midrule
SEGA         & \textbf{1.30}  & \textbf{23.26} & \textbf{29.30} & 52.97 & \textbf{151.08} & \textbf{43.86} & \textbf{1.26}  & \textbf{23.18} & \textbf{29.22} & \textbf{45.73} & \textbf{150.05} & \textbf{51.28} \\
\bottomrule
\end{tabular}
}
\end{table}

%% file: tables/qwen-aesthetic.tex
\begin{table*}[t]
\centering
\scriptsize
\caption{Quantitative comparison on Qwen across all four resolutions on Aesthetic-4K~\cite{zhang2025diffusion}.}
\vspace{5pt}
\label{tab:resolution_all_metrics}
\begin{tabular}{ll cccccccc}
\toprule
Resolution & Method & IR$\uparrow$ & PS$\uparrow$ & HPS$\uparrow$ & CS$\uparrow$ & MSQ$\uparrow$ & CQA$\uparrow$ & FID$\downarrow$ & FID$_p$$\downarrow$ \\
\midrule
\multirow{4}{*}{$2048 \times 4096$}
 & Base       & 0.60 & 21.77 & 0.25 & 28.17 & 45.05 & 0.64 & \underline{156.53} & 60.33 \\
 & DyPE       & \underline{1.07} & \underline{22.20} & \underline{0.27} & 28.60 & \underline{47.44} & \underline{0.65} & 157.86 & 75.97 \\
 & UltraImage & 0.90 & 21.61 & 0.24 & \underline{28.66} & 47.21 & 0.63 & 158.97 & \textbf{41.77} \\
 & SEGA       & \textbf{1.50} & \textbf{23.63} & \textbf{0.30} & \textbf{29.75} & \textbf{50.52} & \textbf{0.72} & \textbf{149.49} & \underline{53.07} \\
\midrule
\multirow{4}{*}{$4096 \times 2048$}
 & Base       & -0.40 & 20.84 & 0.23 & 27.70 & 42.93 & 0.63 & 191.89 & 70.43 \\
 & DyPE       & 0.44 & \underline{21.39} & \underline{0.26} & \underline{28.73} & \underline{48.06} & \underline{0.65} & \underline{159.38} & 81.79 \\
 & UltraImage & \underline{0.46} & 21.04 & 0.23 & 28.21 & 46.11 & 0.63 & 164.06 & \textbf{40.16} \\
 & SEGA       & \textbf{1.27} & \textbf{23.09} & \textbf{0.29} & \textbf{29.94} & \textbf{51.23} & \textbf{0.74} & \textbf{147.51} & \underline{51.10} \\
\midrule
\multirow{4}{*}{$3072 \times 3072$}
 & Base       & 0.37 & \underline{23.17} & 0.25 & 28.03 & 43.07 & 0.65 & 162.39 & 58.61 \\
 & DyPE       & \underline{1.13} & 22.38 & \underline{0.28} & \underline{29.03} & 47.12 & \underline{0.70} & \underline{151.71} & 66.44 \\
 & UltraImage & 1.04 & 22.08 & 0.26 & 29.00 & \underline{47.92} & 0.66 & 153.50 & \textbf{35.80} \\
 & SEGA       & \textbf{1.49} & \textbf{23.67} & \textbf{0.32} & \textbf{29.97} & \textbf{47.95} & \textbf{0.72} & \textbf{150.31} & \underline{46.22} \\
\midrule
\multirow{4}{*}{$4096 \times 4096$}
 & Base       & -0.10 & 20.95 & 0.21 & 27.82 & 28.48 & 0.54 & 174.00 & 71.25 \\
 & DyPE       & \underline{0.97} & \underline{22.04} & 0.27 & \underline{29.08} & \underline{37.39} & \underline{0.63} & \underline{159.78} & 66.00 \\
 & UltraImage & 0.81 & 21.53 & \underline{0.27} & 28.55 & 34.74 & 0.59 & 167.04 & \underline{65.90} \\
 & SEGA       & \textbf{1.51} & \textbf{23.84} & \textbf{0.33} & \textbf{30.12} & \textbf{45.03} & \textbf{0.74} & \textbf{148.26} & \textbf{65.72} \\
\bottomrule
\end{tabular}
\end{table*}

%% file: tables/flux-aesthetic-ablation-4k.tex
\begin{table}[t]
\centering
\scriptsize
\caption{Ablation study on Flux at 4096$\times$4096 resolution on Aesthetic-4K.~\cite{zhang2025diffusion}}
\label{tab:flux_ablation_4k}
\setlength{\tabcolsep}{7pt}
\renewcommand{\arraystretch}{1.0}
\begin{tabular}{@{} l | ccccccc @{}}
\toprule
Method & IR$\uparrow$ & PS$\uparrow$ & HPS$\uparrow$ & CS$\uparrow$ & MSQ$\uparrow$ & CQA$\uparrow$ & FID$\downarrow$ \\
\midrule
NTK + Fixed Scaling        & 0.66 & 22.10 & 0.270 & 28.38 & 42.21 & 0.690 & 162.92 \\
\midrule
SEGA W/ Axis-only          & 1.15 & \textbf{23.05} & \underline{0.290} & \underline{28.86} & 44.08 & 0.701 & \underline{152.98} \\
SEGA W/ Global-only        & 1.13 & \underline{22.89} & 0.286 & 28.81 & 43.55 & 0.671 & 153.58 \\
\midrule
YaRN + SEGA                & 0.94 & 22.21 & 0.280 & 28.23 & 44.36 & 0.707 & 159.80 \\
DyPE + SEGA                & \underline{1.18} & 22.78 & 0.287 & 28.84 & \underline{45.18} & \underline{0.720} & 154.51 \\
\midrule
\textbf{SEGA}              & \textbf{1.26} & 22.35 & \textbf{0.291} & \textbf{29.22} & \textbf{45.72} & \textbf{0.725} & \textbf{150.05} \\
\bottomrule
\end{tabular}
\end{table}

%% file: section/conclusion.tex
\vspace{-6pt}
\section{Conclusion}
\vspace{-6pt}
We presented SEGA, a training-free method for high-resolution extrapolation in DiTs that adapts RoPE components scaling to the spectral structure of the current latent. By making attention scaling frequency-aware across RoPE components, SEGA addresses a key limitation of existing uniform scaling strategies, which often trade off global coherence against fine-detail fidelity. This simple modification requires no retraining or architectural changes, yet consistently improves structure, semantics, and visual quality across resolutions and model architectures. More broadly, frequency-aware attention scaling may also benefit video and other modalities where resolution extrapolation remains challenging. We hope the spectral perspective guidance introduced here motivates further research on modifying attention behavior, particularly for resolution extrapolation, to better unlock the capacity of pretrained generative models.

%% file: section/appendix.tex
\appendix

\section*{\Large Appendix}


\section{Detailed Related Work and Preliminaries}
\label{app-related-work}

\subsection{High-Resolution Image Synthesis}

\paragraph{Training-Based Approaches}
An orthogonal line of work addresses high-resolution synthesis through fine-tuning on curated high-resolution data. Diffusion-4K~\cite{zhang2025diffusion} fine-tunes latent diffusion models on a dedicated 4K dataset using wavelet-based supervision to reinforce high-frequency fidelity, achieving strong perceptual quality at the cost of retraining and reduced architectural generalizability. Latent Wavelet Diffusion (LWD)~\cite{sigillo2025latent} takes a lighter approach, introducing frequency-aware training objectives, including a scale-consistent VAE loss and spatially adaptive denoising supervision guided by wavelet energy maps. While these methods highlight the value of frequency-domain supervision during training, they remain tied to the fine-tuning regime and do not generalize to arbitrary unseen models or resolutions at inference time.

\paragraph{Training-Free Methods: U-Net Architectures}
Training-free high-resolution generation has been studied extensively in U-Net-based latent diffusion models. DemoFusion~\cite{du2024demofusion} extends pretrained models beyond their native resolution using progressive upscaling, skip residuals, and dilated sampling. FreeScale~\cite{qiu2025freescale} introduces scale fusion with selective frequency extraction, FreCaS~\cite{zhang2024frecas} uses frequency-aware cascaded sampling, ScaleCrafter~\cite{he2023scalecrafter} exploits dilated convolutions at inference, DiffuseHigh~\cite{kim2025diffusehigh} incorporates wavelet-domain guidance, and FouriScale~\cite{huang2024fouriscale} applies Fourier-domain frequency rescaling to suppress repetitive patterns. These methods show that high-resolution generation can be improved at inference time, but their mechanisms are closely tied to U-Net-style pipelines with convolutional feature maps, decoder stages, and skip connections. SEGA instead targets RoPE-based diffusion transformers, where resolution extrapolation is governed by attention over expanded latent token grids rather than explicit multi-scale feature hierarchies.

\paragraph{Training-Free Methods: Diffusion Transformers}
Training-free methods for DiT-based high-resolution generation generally fall into two categories: \emph{direct inference} and \emph{multi-stage guidance} approaches. Multi-stage methods condition high-resolution sampling on guidance extracted from a base-resolution generation. I-Max~\cite{du2024imax} uses projected flows derived from native-resolution generation to stabilize coarse structure formation. HiFlow~\cite{bu2025hiflow} extends this idea by constructing a virtual reference flow from the full low-resolution trajectory, providing initialization, direction, and acceleration guidance. ScaleDiff~\cite{koh2025scalediff} follows a similar cascade paradigm, combining upsample--diffuse--denoise refinement with patch-level attention and latent frequency mixing. While these methods provide strong structural priors, they also tie output quality to the fidelity of the base-resolution generation.

\subsection{RoPE-Based Length Extrapolation Methods}
\label{sec:related_rope}

RoPE-based extrapolation is the line of work most closely related to SEGA. As reviewed in Section~\ref{sec:preliminaries}, existing methods modify the RoPE schedule $\theta_d$, the attention scaling, or both. Below, we summarize how these strategies extend to the 2D spatial setting of image generation.

\paragraph{Two-Dimensional Extrapolation Structure.}
For image generation, RoPE is applied axially~\cite{heo2024rotary}, 
with separate rotary schedules for the height and width components 
of each token. Let $s_H = L^{(H)}_{\mathrm{target}} / L^{(H)}_{\mathrm{train}}$ 
and $s_W = L^{(W)}_{\mathrm{target}} / L^{(W)}_{\mathrm{train}}$
denote the per-axis extrapolation ratios. 

\subsubsection{Position Interpolation (PI)}
Position Interpolation~\cite{chen2023extending} rescales positions 
linearly along each axis, $n^{(a)} \mapsto n^{(a)} / s_a$ for 
$a \in \{H, W\}$,
which is equivalent to uniformly contracting all RoPE frequencies to $\theta_d/s_a$. This maps extrapolated positions back into the training range and reduces phase drift at long positions. However, because the same compression is applied to all dimensions, PI treats coarse long-wavelength structure and fine short-wavelength detail identically, which can weaken high-frequency positional sensitivity at large resolutions.

\subsubsection{NTK}
NTK~\cite{bloc97_ntk} instead modifies the RoPE base along each axis. The original 1D rule uses
\begin{equation}
b' = b \cdot s_a^{D/(D-2)},
\qquad
\theta_d' = (b')^{-2(d-1)/D}.
\end{equation}
In our experiments, this correction is too weak for 2D image extrapolation, where at high resolution, the rescaled frequencies fail to provide adequate positional discrimination in attention, leading to blurred or repetitive outputs. We therefore use a stronger variant,
\begin{equation}
b' = b \cdot s_a^{2D/(D-2)},
\end{equation}
which better preserves positional contrast across the expanded 2D token grid. Unlike PI, NTK is dimension-dependent, but it remains a fixed function of $s_a$ and $d$: it does not adapt to the latent content, the sample, or the denoising state.

\subsubsection{YaRN}
YaRN~\cite{peng2024yarn} refines NTK by partitioning 
RoPE dimensions into frequency bands and applying tailored strategies 
to each. Its frequency interpolation uses a smooth ramp function 
$\lambda_d \in [0,1]$ to blend between PI-style interpolation and 
unmodified extrapolation:
\begin{equation}
    \theta_d' = (1 - \lambda_d)\,\frac{\theta_d}{s_a} + \lambda_d\,\theta_d,
\end{equation}
where $\lambda_d = \lambda(r_d)$ is determined by the normalized 
wavelength ratio $r_d = T_d / L_{\text{train}}$, with 
$T_d = 2\pi/\theta_d$ the wavelength of the $d$-th RoPE dimension:
\begin{equation}
    \lambda(r) =
    \begin{cases}
        0,                                   & \text{if } r < \alpha \\
        1,                                   & \text{if } r > \beta  \\
        \dfrac{r - \alpha}{\beta - \alpha},  & \text{otherwise.}
    \end{cases}
\end{equation}

Although YaRN's mixed interpolation-extrapolation strategy is highly effective in the 1D setting of LLMs, we find that it does not transfer well to 2D image generation.
In our experiments, YaRN frequently produces spatial structure collapse and layout confusion like objects appear in inconsistent locations, global composition breaks down, and semantically distinct regions blend together.
We attribute this to YaRN's dimension-selective frequency blending: in a 1D sequence, partially interpolating high-frequency dimensions while extrapolating low-frequency ones is well-motivated by the monotonic positional structure of text.
In 2D images, however, spatial structure is encoded jointly across both axes and across multiple frequency bands simultaneously, and selectively suppressing certain frequency dimensions disrupts the 2D positional geometry in ways that do not arise in the 1D case.
In contrast, NTK, which rescales all dimensions consistently via the base frequency, better preserves both coarse layout and high-level spatial structure in our experiments, making it a more reliable foundation for 2D extrapolation.

YaRN further introduces a global attention temperature correction. 
As discussed in Section~\ref{sec:temperature}, this is written as a 
logit-level factor:
\begin{equation}
    \tau(s) = \bigl(0.1\ln(s)+1\bigr),
\end{equation}
which sharpens attention distributions at extended lengths to counteract 
the entropy collapse that arises when positional offsets grow beyond 
the training range. 
\subsubsection{DyPE}
DyPE~\cite{issachar2025dype} makes RoPE extrapolation timestep-adaptive. Motivated by the coarse-to-fine progression of diffusion sampling, it replaces the fixed extrapolation ratio $s$ with a timestep-dependent schedule $s(t)$ and applies it to standard RoPE corrections. For example, a Dy-NTK variant uses
\begin{equation}
b'(t) = b \cdot s(t)^{D/(D-2)},
\qquad
\theta_d'(t) = b'(t)^{-2(d-1)/D}.
\end{equation}
DyPE is more adaptive than PI, NTK, and YaRN; however, its adaptation is still driven by a predefined timestep schedule rather than by the observed latent of the current sample. SEGA is complementary, it also evolves during denoising, but derives its modulation directly from the current latent's spectral structure.

\subsubsection{UltraImage}

UltraImage~\cite{zhao2025ultraimage} addresses two failure modes in DiT resolution extrapolation: content repetition and quality degradation. For repetition, it identifies a \emph{dominant frequency}, a mid-band RoPE dimension whose spatial period $T_d = 2\pi/\theta_d$ aligns with the training resolution, and applies a recursive correction that reduces this frequency until its period exceeds the extrapolated extent, eliminating periodic tiling artifacts. For quality degradation, it proposes \emph{entropy-guided adaptive attention concentration}: attention entropy $H_i = -\sum_j A_{ij}\log A_{ij}$ is computed per head and used to assign a focus factor that sharpens diffuse local attention while preserving globally concentrated patterns.

UltraImage is closely related to SEGA, as both are motivated by the view that RoPE behavior and attention degradation are central to high-resolution extrapolation in diffusion transformers. However, the two methods differ fundamentally in both diagnosis and mechanism. UltraImage identifies a discrete set of dominant RoPE frequencies whose spatial periods align with the training resolution and corrects them individually via a recursive procedure. Its attention correction is similarly discrete; an entropy score is computed per attention head and used to assign a scalar focus factor, sharpening heads that have become overly diffuse. Both corrections are therefore \emph{sparse} and \emph{binary} in nature. In contrast, SEGA analyzes the full spectral energy distribution of the current latent and uses it to derive a \emph{continuous}, per-dimension scaling pattern that varies across all RoPE dimensions and both image axes. This means that every RoPE dimension receives a scaling that reflects how much spatial variation the latent currently exhibits at the corresponding frequency band, not just whether that dimension happens to coincide with a dominant period.


\begin{figure}[t]
  \centering
  \includegraphics[width=\textwidth]{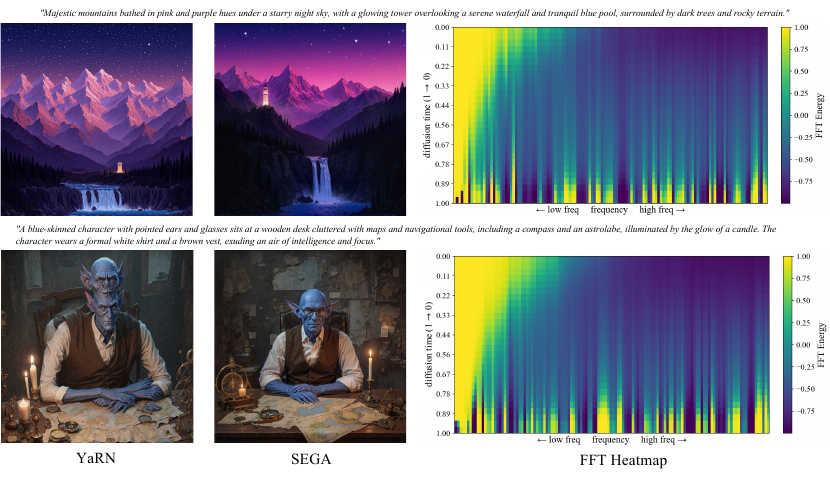}
  \vspace{-15pt}
  \caption{\textbf{Content-Aware Spectral Evolution.} The 2D power spectrum of the intermediate latents across the denoising process for two distinct prompts. The spectral energy distribution varies depending on the image content, demonstrating the necessity of a content-aware approach. Furthermore, the shifting concentration of energy, particularly in low-frequency bands where static over-scaling introduces structural artifacts (as observed in Figure \ref{fig:scaling_tradeoff}), justifies the latent's spectral power as a dynamic guidance signal to adaptively allocate scaling, evaluated on Flux at $4096^2$.}
  \label{fig:fft_heatmap}
\end{figure}

\begin{figure}[t]
  \centering
  \includegraphics[width=\textwidth]{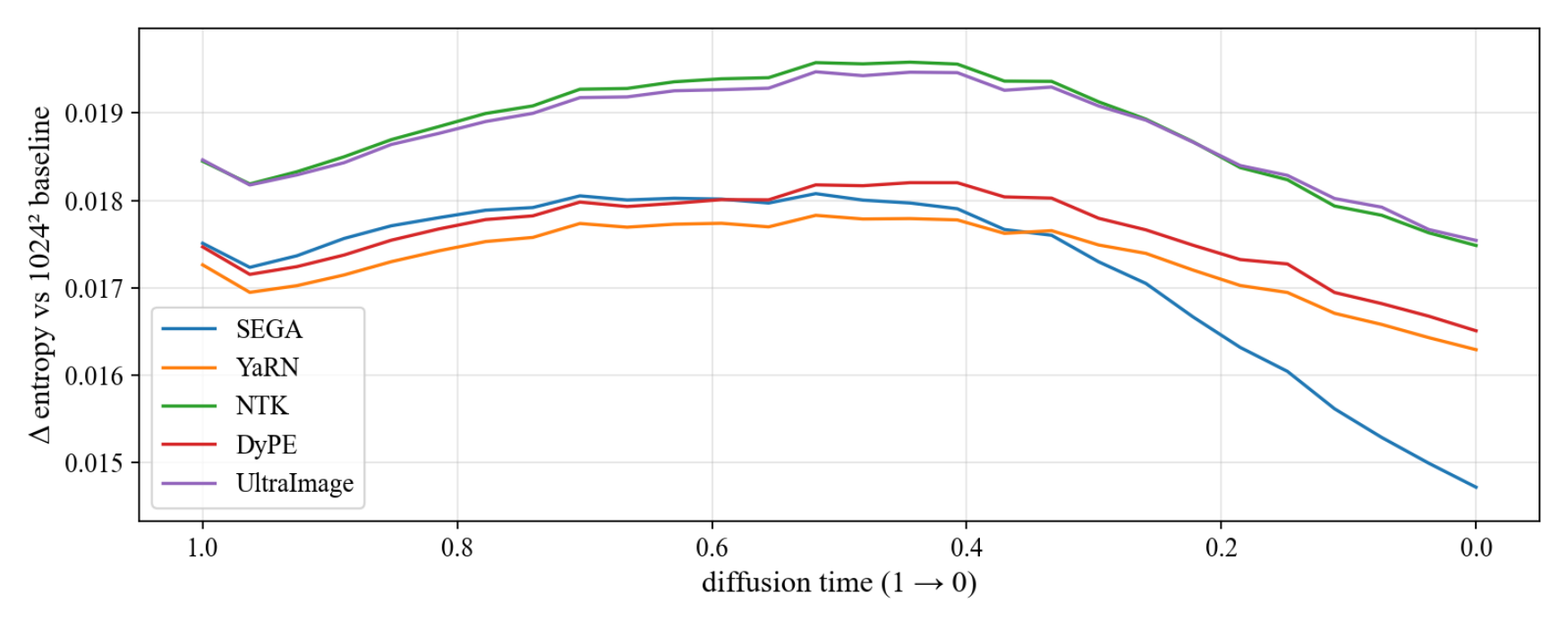}
  \vspace{-15pt}
  \caption{\textbf{Attention Entropy.} The delta of attention entropy value between different methods and the baseline image generated at $1024^2$ resolution on Flux. A smaller difference indicates a closer attention structure to the baseline image generated without any RoPE extrapolation and scaling methods.}
  \label{fig:attention_entropy}
\end{figure}

\begin{figure}[t]
  \centering
  \includegraphics[width=0.85\textwidth]{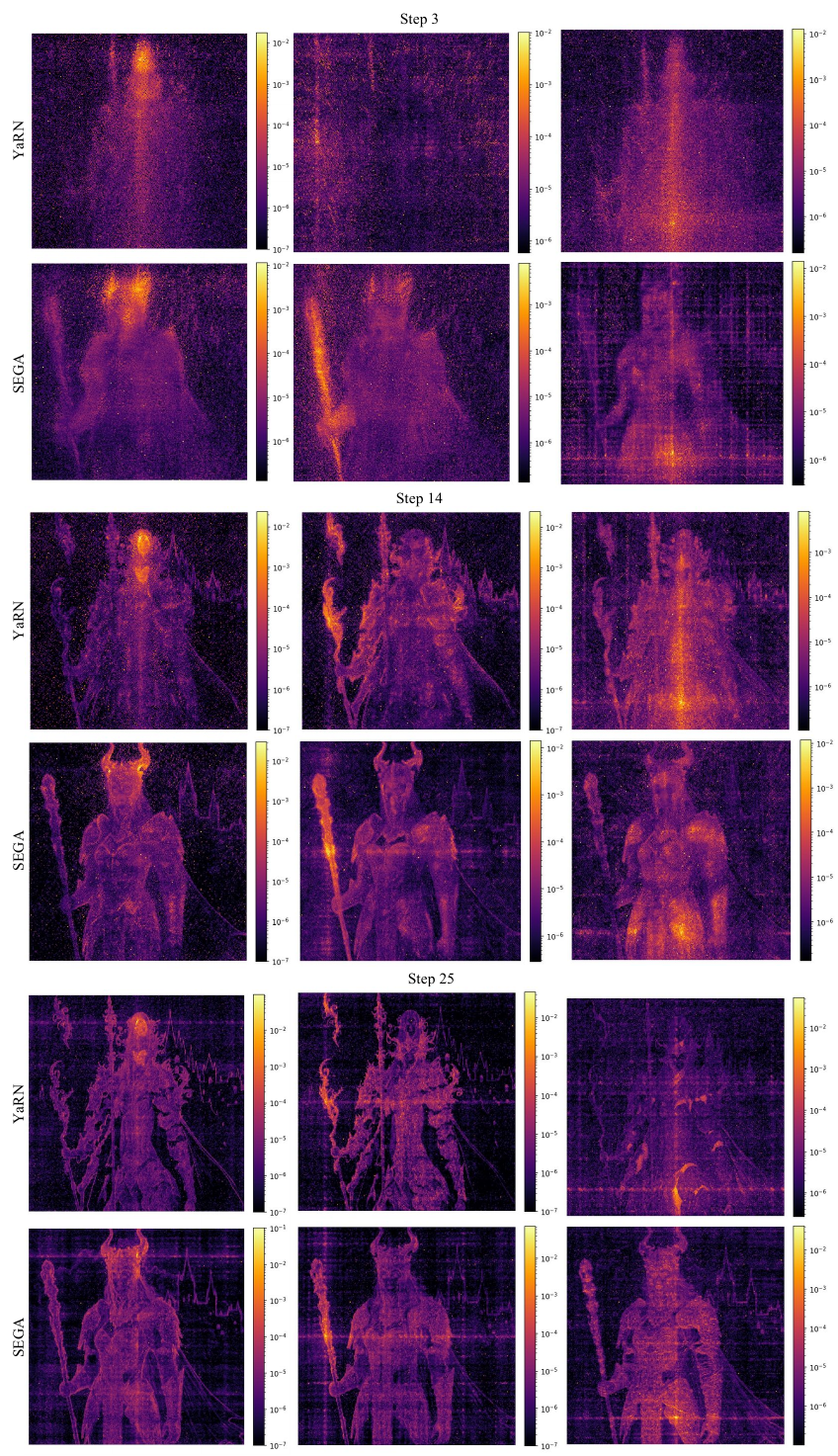}
  \caption{\textbf{Impact on Attention Evolution (Other Tokens).} Further visual comparison of attention maps for the top-center, middle-left, and bottom-center latent tokens in YaRN and SEGA across multiple denoising steps, evaluated on Flux at $4096^2$. \vspace{-10pt}}
  \label{fig:attention_map_full}
\end{figure}


\section{Additional Analysis of Spectral-Energy Guided Attention}
\label{app:analysis}

\subsection{Content-Dependent Spectral Structure of Latent Representations}
\label{app:fft_analysis}

A central premise of SEGA is that the spectral energy distribution of the latent $\mathbf{Z}$ is not fixed. it varies across prompts, 
semantic content, and denoising timesteps, and this variation carries 
meaningful signal about how attention scaling should be allocated across 
RoPE dimensions.
Figure~\ref{fig:fft_heatmap} provides direct empirical support for 
this premise. Each heatmap shows the normalized 2D power spectrum of the intermediate latent tokens across the denoising trajectory, from pure noise (bottom, $t \approx 1$) to the final generated image (top, $t \approx 0$), for two prompts with markedly different visual characteristics: a landscape scene with large-scale spatial structure (top heatmap) and a portrait scene with dense local texture and fine detail (bottom heatmap).

Two observations are immediately apparent. First, the spectral energy distributions differ between the two prompts. The landscape latent develops a broader spread of energy into mid- and high-frequency bands, reflecting its detailed textures (water, foliage, rocks), whereas the portrait latent concentrates more sharply in the low-frequency region, consistent with its smoother large-scale structure. This inter-prompt variability directly motivates the content-aware design of SEGA: a fixed, globally-defined RoPE scaling, as used by YaRN and DyPE, cannot simultaneously be optimal for both spectral profiles. Applying the same frequency schedule to both prompts inevitably over-scales some bands and under-scales others, depending on where the image's structural energy actually resides.

Second, within each prompt, the spectral energy distribution evolves across the denoising trajectory. Early in denoising (bottom of each heatmap), when the latent is dominated by noise, the spectrum is highly variable across frequency bins, with no clear concentration in some specific bands. As denoising proceeds, low-frequency components emerge first and become increasingly dominant, establishing the coarse global structure of the image, while the high-frequency region remains comparatively low-energy, with its residual content varying subtly depending on the image's texture complexity. By the end of the trajectory (top of each heatmap), energy is sharply concentrated in the low-frequency region, with a smaller but content-dependent contribution in the higher bands. This temporal evolution, from an irregular noise-dominated spectrum to a structured one shaped by image content, further motivates SEGA's design of recomputing the spectral profile at each denoising step rather than fixing it at the start of sampling.

\subsection{Attention Entropy Analysis}
\label{app:entropy_analysis}

A useful signal of extrapolation quality is how closely the attention structure at high resolution resembles that of the model within its training distribution.
When attention entropy deviates substantially from the baseline,  the model's capacity to allocate focus appropriately is compromised, either through excessive diffusion of attention mass (high-entropy, diluted attention) or through concentration on a small number of tokens (low-entropy, collapsed 
attention).
DyPE~\cite{issachar2025dype} has shown that resolution extrapolation typically induces a shift in attention entropy relative to the training distribution, and that methods which minimize this shift tend to produce higher-quality outputs.

Figure~\ref{fig:attention_entropy} reports the delta attention entropy, the difference in mean attention entropy between each extrapolation 
method and the baseline Flux model operating at its native 
$1024^2$ resolution as a function of the denoising timestep, 
averaged across different seeds, prompts from Aesthetic-4K~\cite{zhang2025diffusion}, and all attention layers and heads.
All methods are evaluated at $4096^2$ resolution.

\subsection{Additional Attention Evolution Results}
To further illustrate how SEGA's content-aware scaling affects 
attention behavior at the token level, Figure~\ref{fig:attention_map_full} 
extends the attention map analysis from Section~\ref{main-analysis} to additional 
spatial locations, specifically the top-center, middle-left, and 
bottom-center latent tokens, comparing YaRN and SEGA across multiple 
denoising steps at $4096^2$ resolution, consistent with the findings reported for the center token in Section~\ref{main-analysis}. The consistency of this behavior across spatially diverse token positions, covering the corners, edges, and interior of the latent grid, confirms that SEGA's improvements are not localized to a particular region of the image but reflect a global improvement in attention structure throughout the high-resolution token grid.


\section{Additional Implementation Details}
All image generation experiments were conducted using the Flux and Qwen diffusion transformer architectures. For the Flux model, we specifically utilized the \texttt{dev.Krea} checkpoint. To maintain high numerical precision without incurring unnecessary memory overhead, all model weights and latent activations were cast to \texttt{bfloat16}. The experiments, including both standard generation and high-resolution extrapolation, were executed on NVIDIA H100 GPUs. Because SEGA operates entirely at inference time and requires no parameter updates, we did not employ any training or fine-tuning infrastructure. We followed the standard inference settings provided by the official model implementations of Flux and Qwen, using their default samplers, number of denoising steps, and guidance scales. SEGA was applied at every denoising step throughout the entire trajectory, with no warmup, scheduling, or step-dependent gating beyond what is induced naturally by the spectral flatness factor.


\section{Limitation and Discussion}

While SEGA enables stable high-resolution synthesis well beyond the native training regime, it has several limitations. First, SEGA modulates the magnitude of rotary embeddings but does not extend RoPE's positional range; it is therefore composed with an underlying length-extrapolation method (NTK in our experiments) and partially inherits its structural priors. Second, although SEGA can scale up to $8192^2$, perceptual quality continues to degrade at the most extreme extrapolation factors, where the limitation is the model's intrinsic capacity rather than the positional encoding alone. Third, while SEGA itself is computationally negligible, generating multi-megapixel images remains expensive: the underlying transformer's attention cost grows quadratically with the number of tokens, making ultra-high-resolution synthesis demanding regardless of which extrapolation method is used.  More broadly, SEGA shows that the latent's spectral structure can serve as a useful signal for guiding RoPE scaling at inference time, and we hope the coupling it reveals between RoPE dimensions and spatial frequencies inspires future work on inference-time adaptation of pretrained generative models.


\section{Additional Quantitative Results}
\label{additional-results}

\input{tables/mix-aesthetic-main}
\input{tables/zero-shot}
\input{tables/aesthetic-5k}

\input{tables/aesthetic-6k}

\subsection{Generalization to Alternative Backbones}
\label{app:sdxl_results}

To assess the generalizability of SEGA beyond Flux-based models, Table~\ref{tab:flux_extended} reports quantitative results on an alternative backbone across four high-resolution settings on Aesthetic-4K~\cite{zhang2025diffusion}. We compare against a broad set of state-of-the-art baselines, including methods built on SDXL~\cite{podell2023sdxl}, as well as Diffusion-4K, which relies on model fine-tuning. The baselines include fine-tuning (Diffusion-4K) and multi-stage guidance (DemoFusion, FreCas, FreeScale, DiffuseHigh). SEGA consistently achieves the best or second-best performance 
across the majority of metrics and resolution settings, demonstrating 
that its spectral-energy-guided scaling transfers effectively across 
different model architectures without any architecture-specific 
tuning.

\subsection{Zero-Shot Benchmark}
\label{app:zeroshot}

A potential concern with evaluating on Aesthetic-4K~\cite{zhang2025diffusion} 
is that some models, particularly those with large-scale pretraining 
data,  may have encountered images from this dataset during training, 
which could favor their performance on distribution-specific metrics.
To mitigate this risk and assess generalization, we construct a dedicated \textbf{\textit{zero-shot}} 
benchmark.

Specifically, we use an LLM to generate 200 curated, 
high-detail prompts covering a diverse range of scenes, lighting 
conditions, subjects, artistic styles, and compositional 
structures, with care taken to minimize overlap with the Aesthetic-4K dataset.
This benchmark is designed to evaluate whether performance 
differences observed on Aesthetic-4K reflect genuine generalization 
capability or are partly attributable to dataset familiarity.
We additionally include Nano Banana 2~\cite{team2023gemini}, a closed-source proprietary model, in this evaluation as a reference point for the performance 
ceiling achievable by large-scale commercial systems.

Table~\ref{tab:zeroshot_resolution} reports results on this 
zero-shot benchmark at $4096^2$ resolution.
SEGA achieves the best performance across all metrics on both the 
Flux and Qwen backbones. Notably, SEGA on the Qwen backbone achieves an ImageReward score of 1.58 and a PickScore of 23.86, approaching and in some metrics 
matching or even better than the performance of Nano Banana 2 (IR: 1.37, PS: 23.43), 
which represents a strong closed-source commercial baseline.

\subsection{Extreme Resolution: $\mathbf{5120^2}$ and $\mathbf{6144^2}$}
\label{app:extreme_resolution}

Tables~\ref{tab:5k_resolution} and~\ref{tab:6k_resolution} extend 
the main evaluation to extreme resolutions of $5120^2$ 
and $6144^2$, corresponding to approximately 26 and 38 
million pixels respectively, resolutions that represent a $25\times$ 
and $36\times$ area extrapolation factor beyond the $1024^2$ 
training resolution of Flux.
Due to the time and cost of generation at these scales, we evaluate on a randomly selected subset of 20 prompt--image pairs from Aesthetic-4K. The results show that SEGA remains substantially more consistent as resolution increases, while competing methods degrade significantly under stronger extrapolation. Its superiority is most pronounced at ultra-high resolutions, where it achieves the strongest overall performance while better preserving structural coherence and semantic fidelity.

\subsection{Comparison with Training-Based Methods}
\label{app:training_based}
 
SEGA is training-free, so the inference-time methods in the main paper
are its most direct baselines. For a broader comparison, we also evaluate two training-based methods
with publicly released checkpoints: UltraPixel~\cite{ren2024ultrapixel},
which builds on a cascaded diffusion model, and
URAE~\cite{yu2025urae}, which fine-tunes Flux for 2K and 4K generation.  Because URAE uses the same base model as SEGA, we also apply SEGA on top of it (URAE + SEGA) to test whether SEGA's
inference-time scaling adds to the gains of fine-tuning.
All methods are evaluated on the full Aesthetic-4K benchmark at
$2048 \times 4096$, $4096 \times 2048$, and $4096^2$. In addition to the
metrics in Sec.~\ref{sec:exps}, 
we report the no-reference quality metric
ARNIQA~\cite{agnolucci2024arniqa}.
 
Table~\ref{tab:training_based} shows the results. Without any training,
SEGA scores higher than both UltraPixel and URAE on CLIP-IQA, MUSIQ,
ARNIQA, and FID at all three resolutions, and on ImageReward at the two
non-square resolutions. The trained methods score higher on CLIP Score
and HPS at all three resolutions, and URAE has a higher ImageReward at
$4096^2$. Applying SEGA to URAE improves over URAE alone on every metric
at every resolution, which indicates that SEGA's inference-time scaling
is complementary to high-resolution fine-tuning. URAE + SEGA gives the
best ImageReward, HPS, and CLIP-IQA at all three resolutions, while SEGA
alone remains best on ARNIQA at all resolutions, on FID at
$2048 \times 4096$ and $4096^2$, and on MUSIQ at $4096^2$.
 
\begin{table}[h]
\centering
\caption{Comparison with training-based methods on the full
Aesthetic-4K benchmark. UltraPixel and URAE are trained for
high-resolution generation; SEGA is training-free and applied to Flux;
URAE + SEGA applies SEGA to URAE. CQA: CLIP-IQA. Best results per
resolution are in \textbf{bold}.}
\label{tab:training_based}
\small
\begin{tabular}{c|l|ccccccc}
\toprule
Res. & Method & IR $\uparrow$ & CS $\uparrow$ & HPS $\uparrow$ & CQA $\uparrow$ & MSQ $\uparrow$ & ARNIQA $\uparrow$ & FID $\downarrow$ \\
\midrule
\multirow{4}{*}{\rotatebox[origin=c]{90}{\footnotesize $2048{\times}4096$}}
    & UltraPixel  & 1.10          & \textbf{29.73} & 0.2853          & 0.6771          & 47.14          & 0.4872          & 156.30 \\
    & URAE        & 1.18          & 29.21          & 0.2846          & 0.6859          & 51.01          & 0.5149          & 153.89 \\
    & SEGA        & 1.21          & 29.18          & 0.2810          & 0.7008          & 53.64          & \textbf{0.6061} & \textbf{151.93} \\
    & URAE + SEGA & \textbf{1.31} & 29.44          & \textbf{0.2882} & \textbf{0.7228} & \textbf{53.80} & 0.5616          & 152.16 \\
\midrule
\multirow{4}{*}{\rotatebox[origin=c]{90}{\footnotesize $4096{\times}2048$}}
    & UltraPixel  & 0.85          & 29.17          & 0.2853          & 0.7003          & 46.42          & 0.4735          & 157.72 \\
    & URAE        & 0.81          & 29.05          & 0.2859          & 0.7060          & 50.99          & 0.5153          & 154.38 \\
    & SEGA        & 0.86          & 28.99          & 0.2836          & 0.7251          & 52.86          & \textbf{0.6122} & 153.10 \\
    & URAE + SEGA & \textbf{0.88} & \textbf{29.23} & \textbf{0.2905} & \textbf{0.7426} & \textbf{53.25} & 0.5599          & \textbf{152.83} \\
\midrule
\multirow{4}{*}{\rotatebox[origin=c]{90}{\footnotesize $4096{\times}4096$}}
    & UltraPixel  & 1.12          & 29.60          & 0.2961          & 0.7193          & 36.22          & 0.4229          & 152.93 \\
    & URAE        & 1.29          & 29.46          & 0.2970          & 0.7204          & 42.17          & 0.4846          & 154.98 \\
    & SEGA        & 1.26          & 29.22          & 0.2913          & 0.7245          & \textbf{45.72} & \textbf{0.5751} & \textbf{150.05} \\
    & URAE + SEGA & \textbf{1.32} & \textbf{29.90} & \textbf{0.3023} & \textbf{0.7463} & 44.56          & 0.5590          & 150.39 \\
\bottomrule
\end{tabular}
\end{table}


\section{Additional Ablation}
\label{app:baseline_ablation}

\subsection{Choice of Baseline Scaling $m_{\text{ref}}$}

The reference scale $m_{\text{ref}}$ in Eq.~\ref{eq:m_ref} sets the anchor magnitude of the rotary scaling shared across all RoPE dimensions. As discussed in Sec.~\ref{sec:allocation}, $m_{\text{ref}}$ is a function of the resolution ratio $s = R_{\text{target}} / R_{\text{train}}$ between target and training images. We consider two common formulations for this design choice:
\begin{equation}
    m_{\text{ref}}^{\text{power}} = s^{\kappa}, 
    \qquad 
    m_{\text{ref}}^{\text{log}} = 1 + \kappa \log s,
\end{equation}
where $\kappa > 0$ is a small exponent (we use $\kappa = 0.08$ in all reported experiments). Both formulations reduce to $m_{\text{ref}} = 1$ at $s = 1$ (no extrapolation) and grow monotonically with $s$. The two forms behave similarly in the moderate-extrapolation regime ($s \approx 1$--$2$), but diverge as $s$ grows.

\paragraph{Why the choice matters at high $s$.}
As the target resolution increases, the token grid grows substantially, making positional offsets harder to discriminate even with RoPE extrapolation. Attention therefore becomes increasingly prone to dilution at large extrapolation factors. A larger $m_{\text{ref}}$ acts as a stronger anchor for positional discrimination, sharpening attention more aggressively to compensate for the expanded grid. Empirically, we find that ultra-high-resolution generation (e.g., $5120^2$ or $6144^2$) requires a stronger anchor than moderate extrapolation, and the power-law form provides this naturally because $s^{\kappa}$ grows faster than $1 + \kappa \log s$. Table~\ref{tab:mbase_values} illustrates this divergence: the two forms are nearly identical at small $s$, but the power-law value becomes meaningfully larger as $s$ increases.

\begin{table}[h]
\centering
\caption{Values of $m_{\text{ref}}$ produced by the two formulations as a function of the resolution ratio $s$. Computed with $\kappa = 0.08$. The power-law form grows faster at large $s$, providing a stronger positional-discrimination anchor at extreme extrapolation factors.}
\label{tab:mbase_values}
\begin{tabular}{c|cccccc}
\toprule
$s$ & 1 & 2 & 4 & 8 & 16 & 32 \\
\midrule
$m_{\text{ref}}^{\text{power}}$ & 1.000 & 1.057 & 1.118 & 1.182 & 1.249 & 1.320 \\
$m_{\text{ref}}^{\text{log}}$ & 1.000 & 1.055 & 1.111 & 1.166 & 1.222 & 1.277 \\
\bottomrule
\end{tabular}
\end{table}

\paragraph{Empirical comparison.}
We compare the two formulations under identical SEGA settings on FLUX at $4096^2$, $5120^2$, and $6144^2$. Table~\ref{tab:mbase_ablation} shows that the two forms perform similarly at $4096^2$, with a small but consistent advantage for the power-law variant. The gap widens at $5120^2$ and remains clear at $6144^2$, where the power-law form yields lower FID and stronger alignment across most metrics. Overall, the power-law baseline extrapolates more stably as resolution increases, matching the trend in Table~\ref{tab:mbase_values}. We therefore adopt the power-law form, which grows faster than a logarithm while remaining more moderate than a linear scaling.

\begin{table}[h]
\centering
\caption{Comparison of power-law and logarithmic forms for $m_{\text{ref}}$ on Flux. SEGA hyperparameters are held constant at $\gamma = 1.5$, $\kappa = 0.08$. Best results per resolution are in \textbf{bold}.}
\label{tab:mbase_ablation}
\small
\begin{tabular}{l|l|cccccc}
\toprule
Resolution & $m_{\text{ref}}$ form & CS $\uparrow$ & ImageReward $\uparrow$ & HPS $\uparrow$ & PickScore $\uparrow$ & FID $\downarrow$ & MUSIQ $\uparrow$ \\
\midrule
\multirow{2}{*}{$4096 \times 4096$} 
    & logarithmic    & 28.46 & 1.23 & 0.29 & 23.10 & 150.16 & 44.65 \\
    & power-law      & \textbf{29.22} & \textbf{1.26} & \textbf{0.29} & \textbf{23.18} & \textbf{150.05} & \textbf{45.73} \\
\midrule
\multirow{2}{*}{$5120 \times 5120$}
    & logarithmic    & 28.41 & 0.80 & 0.28 & 22.93 & 265.01 & 46.36 \\
    & power-law      & \textbf{28.92} & \textbf{1.13} & \textbf{0.29} & \textbf{23.22} & \textbf{221.99} & 40.64 \\
\midrule
\multirow{2}{*}{$6144 \times 6144$}
    & logarithmic    & 27.54 & 0.67 & 0.27 & 22.37 & 269.58 & 42.92 \\
    & power-law      & \textbf{27.92} & \textbf{0.75} & \textbf{0.27} & \textbf{22.47} & \textbf{232.18} & \textbf{43.43} \\
\bottomrule
\end{tabular}
\end{table}

\subsection{Radial Bin Selection for Spectral Flatness}
\label{app:radial_bins_ablation}

Spectral flatness (Sec.~\ref{sec:allocation}) is computed from the isotropic radial
energy profile of the packed latent grid, using
\begin{equation}
    n_{\text{bins}} = \left\lfloor \frac{\max(H, W)}{2} \right\rfloor,
\end{equation}
where $H$ and $W$ are the height and width of the packed latent grid. The
number of bins is therefore not fixed but scales with the target resolution.
This choice matches the Nyquist limit, so that each radial bin corresponds to
roughly one integer frequency radius and the profile retains near-native
radial frequency resolution at every image size (e.g., $n_{\text{bins}} = 128$
at $4096^2$).

Using too few bins merges distinct frequency bands, while using too many
leaves bins sparsely populated and makes the profile noisy; we therefore
expect performance to be stable over a reasonable range around the default.
To verify this, we vary $n_{\text{bins}}$ while holding all other SEGA
settings fixed, evaluating on 20 randomly selected prompts at $4096^2$. As
shown in Table~\ref{tab:radial_bins_ablation}, the method is largely
insensitive to this choice: all metrics change only marginally between 28 and
128 bins, and the default of 128 achieves the best CS, HPS, and Patch FID.
Increasing beyond the Nyquist-matched value to 256 bins leads to a slight
drop across all metrics, most visibly in ImageReward.

\begin{table}[h]
\centering
\caption{Effect of the number of radial bins $n_{\text{bins}}$ used to compute
spectral flatness, evaluated on FLUX at $4096 \times 4096$ over 20 prompts randomly
selected from Aesthetic-4K. All other SEGA hyperparameters are held fixed. Best results
are in \textbf{bold}.}
\label{tab:radial_bins_ablation}
\small
\begin{tabular}{l|ccccc}
\toprule
$n_{\text{bins}}$ & CS $\uparrow$ & ImageReward $\uparrow$ & HPS $\uparrow$ & Patch FID $\downarrow$ & MUSIQ $\uparrow$ \\
\midrule
28            & 28.27          & \textbf{0.999} & 0.2783          & 162.7          & 46.16          \\
64            & \textbf{28.31} & 0.995          & 0.2786          & 161.0          & \textbf{46.21} \\
128 (default) & \textbf{28.31} & 0.998          & \textbf{0.2791} & \textbf{160.6} & 46.20          \\
256           & 28.10          & 0.962          & 0.2788          & 161.8          & 46.01          \\
\bottomrule
\end{tabular}
\end{table}

\subsection{Per-Image Adaptation versus a Shared Averaged Schedule}
\label{app:averaged_schedule_ablation}
 
SEGA's improvement over fixed scaling (Table~\ref{tab:flux_ablation_4k})
could come from a per-dimension, per-step scaling pattern shared across
images, or from adapting that pattern to each image. To separate the two, we build an \emph{averaged schedule}. We run SEGA on
Flux at $4096^2$ for 20 prompts randomly selected from Aesthetic-4K and
record the scale $m_d^{(a)}$ that SEGA assigns at each denoising step.
We then average these scales over the 20 images, keeping a separate
value for each axis, RoPE dimension, and step. This schedule is then
frozen and applied to every image without computing the spectral
profile; all other settings are unchanged.
 
Table~\ref{tab:averaged_schedule} reports results on the full
Aesthetic-4K benchmark and on the same 20 randomly selected Aesthetic-4K
prompts used to build the schedule, a setting that favors the averaged
schedule. SEGA is better on all six
metrics in both settings. The gaps are small: the averaged schedule
recovers most of SEGA's gain over fixed scaling (e.g., ImageReward
$0.66 \rightarrow 1.20$, versus $1.26$ for SEGA), so much of the benefit
comes from the shared structure produced by SEGA's spectral analysis,
and per-image adaptation adds a smaller but consistent improvement.
Unlike SEGA, the averaged schedule must also be recorded separately for
each model, resolution, and number of sampling steps.
 
\begin{table}[h]
\centering
\caption{SEGA versus an averaged schedule on Flux at $4096 \times 4096$.
\emph{Full benchmark} uses all Aesthetic-4K prompts; \emph{Subset}
uses the 20 prompts randomly selected from Aesthetic-4K to build the
averaged schedule. FID on these 20 images has high variance. Best results per
evaluation set are in \textbf{bold}.}
\label{tab:averaged_schedule}
\small
\begin{tabular}{l|l|cccccc}
\toprule
Prompts & Method & CS $\uparrow$ & IR $\uparrow$ & HPS $\uparrow$ & PS $\uparrow$ & FID $\downarrow$ & MSQ $\uparrow$ \\
\midrule
\multirow{2}{*}{Full benchmark}
    & Averaged schedule & 28.98 & 1.20 & 0.286 & 23.12 & 152.36 & 44.85 \\
    & SEGA & \textbf{29.22} & \textbf{1.26} & \textbf{0.291} & \textbf{23.18} & \textbf{150.05} & \textbf{45.73} \\
\midrule
\multirow{2}{*}{Subset (20)}
    & Averaged schedule & 29.06 & 1.08 & 0.282 & 23.05 & 243.79 & 47.15 \\
    & SEGA & \textbf{29.40} & \textbf{1.12} & \textbf{0.289} & \textbf{23.10} & \textbf{231.84} & \textbf{47.55} \\
\bottomrule
\end{tabular}
\end{table}
 
\paragraph{How SEGA's scales differ across images.}
The averaged schedule keeps the part of SEGA's scaling that is shared
across images and removes the part that differs from image to image.
To see what is removed, we measure how much $m_d^{(a)}$ varies across
the 20 images at each dimension and step. Three observations stand out.
First, the variation follows image formation: it is near zero at the
first step, where every latent is pure noise and $\sigma \approx 0$,
and appears once structure starts to form (mean standard deviation of
0.016 by step 3, then 0.013--0.015 until the end). Second, it is not
uniform across dimensions: it is largest for dimensions 23 and 24, and
the most variable dimension changes during denoising (27 early, 24 in
the middle, 21 at the end). Third, for about 20\% of dimension--step
pairs, SEGA scales above $m_{\text{ref}} = 1.117$ for some images and
below it for others; for example, dimension 24 at step 15 ranges from
1.000 to 1.153 across images. The averaged schedule assigns a single
value to each pair, so it cannot strengthen the scaling of a dimension
for one image while weakening it for another. These image-specific
differences are the only difference between SEGA and the averaged
schedule, so they account for the gap in Table~\ref{tab:averaged_schedule}.

\subsection{Channel Averaging in the Spectral Estimate}
\label{app:channel_averaging_ablation}
 
SEGA applies one scaling profile to all channels, so it needs a single
estimate of the latent's spatial structure. As described in
Sec.~\ref{sec:analysis}, 
it averages the hidden states over channels before the FFT and uses the
power spectrum
\begin{equation}
    \Bigl|\mathcal{F}_{\text{2D}}\Bigl[\tfrac{1}{C}\textstyle\sum_{c=1}^{C} \mathbf{x}_c\Bigr]\Bigr|^2,
\end{equation}
where $\mathbf{x}_c$ is the $c$-th channel arranged on the $H \times W$
token grid. Because the FFT is linear, structure that has opposite signs
in different channels can cancel in this average. An alternative that
avoids this cancellation averages the power spectra after the FFT,
\begin{equation}
    \frac{1}{|S|} \sum_{c \in S} \bigl|\mathcal{F}_{\text{2D}}[\mathbf{x}_c]\bigr|^2,
\end{equation}
where $S$ is a set of channels. We compare the two on Flux at $4096^2$
over 20 prompts randomly selected from Aesthetic-4K, using random
subsets of $|S| = 256$ and $1024$ channels and all $C = 3072$ channels.
All other SEGA settings are unchanged.
 
Table~\ref{tab:channel_averaging_ablation} reports how much each
alternative changes $\sigma$ and the per-dimension scales relative to
channel averaging, together with image quality and the memory used by
the spectral estimate. Averaging after the FFT changes $\sigma$ by about
0.04--0.05 and the scales by about 1.5\% on average. The effect on
quality is mixed: no channel-wise variant is better than channel
averaging on all metrics, and using all channels gives the lowest
ImageReward and CLIP Score. The channel-wise estimators also need 2 to
7 times more transient memory. We therefore keep channel averaging.
 
\begin{table}[h]
\centering
\caption{Spectral estimators for SEGA on Flux at $4096 \times 4096$,
evaluated on 20 prompts randomly selected from Aesthetic-4K. Channel
averaging (default) computes the power spectrum of the channel-averaged
latent; the other rows average per-channel power spectra over $|S|$
channels. $|\Delta\sigma|$: mean absolute change in $\sigma$;
$\Delta m$: mean relative change in $m_d^{(a)}$, both with respect to
channel averaging. Memory: transient memory, in MB, used by the
spectral estimate. 
Best results are in \textbf{bold}.}
\label{tab:channel_averaging_ablation}
\small
\begin{tabular}{l|cc|cccc|c}
\toprule
Estimator & $|\Delta\sigma|$ & $\Delta m$ (\%) & CS $\uparrow$ & IR $\uparrow$ & HPS $\uparrow$ & MSQ $\uparrow$ & Memory $\downarrow$ \\
\midrule
Channel averaging   & --    & --   & \textbf{28.41} & 1.060          & \textbf{0.2875} & 49.26          & \textbf{768} \\
$|S| = 256$         & 0.048 & 1.49 & 28.31          & 1.056          & 0.2858          & 49.29          & 1{,}536 \\
$|S| = 1024$        & 0.043 & 1.46 & 28.24          & \textbf{1.070} & 0.2864          & 49.47          & 1{,}792 \\
$|S| = 3072$ (all)  & 0.042 & 1.45 & 28.19          & 1.026          & 0.2861          & \textbf{49.57} & 5{,}376 \\
\bottomrule
\end{tabular}
\end{table}

\subsection{Sensitivity to $\kappa$ and $\gamma$}
\label{app:hyperparameter_sensitivity}
 
SEGA has two important scalar hyperparameters. The exponent $\kappa$ in
Eq.~\ref{eq:m_ref} sets how fast the reference scale $m_{\text{ref}}$
grows with the extrapolation ratio. The exponent $\gamma$ in
Eq.~\ref{eq:sigma} 
sets how quickly the amplitude factor $\sigma$ rises as the spectrum
departs from flatness; since $\mathrm{SF} \in (0, 1]$, a larger $\gamma$
gives a larger $\sigma$ for the same flatness, so the per-dimension
correction is applied more strongly. We selected both on Flux at
$4096^2$, using quantitative results on 20 prompts randomly selected
from Aesthetic-4K together with visual inspection, and use the same
values ($\kappa = 0.08$, $\gamma = 1.5$) for all models and resolutions.
Table~\ref{tab:hyperparameter_sensitivity} varies each parameter while
holding the other at its default.
 
\paragraph{$\kappa$.}
Smaller values give weaker attention sharpening: $\kappa = 0.04$ and
$0.06$ score lower than the default on ImageReward, CLIP Score, HPS, and
MUSIQ, although slightly better on FID$_p$. Increasing $\kappa$ to
$0.10$ or $0.12$ raises ImageReward slightly ($1.004$ and $1.006$ versus
$0.998$) but increases FID$_p$ sharply ($177.5$ and $216.8$ versus
$160.6$), and from $\kappa = 0.14$ all metrics are worse than the
default. $\kappa = 0.08$ gives the best CLIP Score, HPS, and MUSIQ and
the best trade-off across metrics in this sweep. Because
$m_{\text{ref}} = s^{\kappa}$, a fixed $\kappa$ still gives a larger
reference scale at higher resolutions (Table~\ref{tab:mbase_values}).
 
\paragraph{$\gamma$.}
The metrics stay within a narrow range from $\gamma = 1.0$ to $2.0$
(ImageReward $0.974$--$1.031$, CLIP Score $28.31$--$28.90$), so SEGA
does not depend on a precise choice of $\gamma$. The default
$\gamma = 1.5$ gives the best value on every metric; the other settings
show no consistent trend toward either end of the range.
 
\begin{table}[h]
\centering
\caption{Sensitivity to $\kappa$ and $\gamma$ on Flux at
$4096 \times 4096$, evaluated on 20 prompts randomly selected from
Aesthetic-4K. Each block varies one parameter with the other at its
default ($\kappa = 0.08$, $\gamma = 1.5$). FID$_p$ on 20 images has high
variance. Best results per block are in \textbf{bold}.}
\label{tab:hyperparameter_sensitivity}
\small
\begin{tabular}{l|l|ccccc}
\toprule
Parameter & Value & CS $\uparrow$ & IR $\uparrow$ & HPS $\uparrow$ & MSQ $\uparrow$ & FID$_p$ $\downarrow$ \\
\midrule
\multirow{7}{*}{$\kappa$}
    & 0.04           & 28.49          & 0.936          & 0.2769          & 43.20          & \textbf{157.8} \\
    & 0.06           & 28.55          & 0.947          & 0.2789          & 45.05          & 158.5 \\
    & 0.08 (default) & \textbf{29.31} & 0.998          & \textbf{0.2797} & \textbf{45.85} & 160.6 \\ 
    & 0.10           & 29.17          & 1.004          & 0.2794          & 45.33          & 177.5 \\
    & 0.12           & 29.11          & \textbf{1.006} & 0.2788          & 45.59          & 216.8 \\
    & 0.14           & 28.69          & 0.997          & 0.2737          & 39.04          & 296.9 \\
    & 0.16           & 28.08          & 0.982          & 0.2684          & 34.35          & 320.5 \\
\midrule
\multirow{5}{*}{$\gamma$}
    & 1.00           & 28.43          & 0.977          & 0.2788          & 45.30          & 161.6 \\
    & 1.25           & 28.62          & 1.015          & 0.2786          & 46.18          & 173.9 \\
    & 1.50 (default) & \textbf{28.90} & \textbf{1.031} & \textbf{0.2817} & \textbf{46.53} & \textbf{160.6} \\ 
    & 1.75           & 28.31          & 0.998          & 0.2791          & 45.85          & 163.0 \\
    & 2.00           & 28.56          & 0.974          & 0.2792          & 45.83          & 170.8 \\
\bottomrule
\end{tabular}
\end{table}


\section{Additional Qualitative Results}

\begin{figure}[t]
  \centering
  \includegraphics[width=\textwidth]{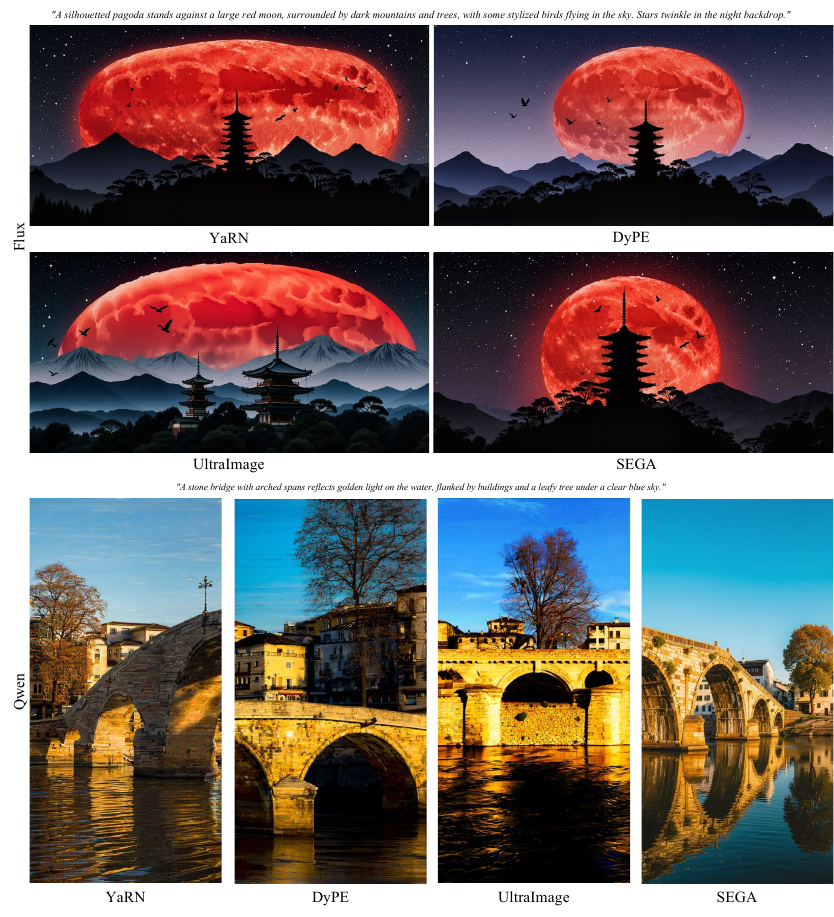}
  \vspace{-15pt}
  \caption{\textbf{Qualitative comparison (non-square resolutions).} Results on two non-square resolutions ($2048 \times 4096$ and $4096 \times 2048$) on Qwen and Flux show that SEGA's ability to preserve the shape of contents in different aspect ratio.}
  \label{fig:qualitative-horiz-vert}
\end{figure}

\begin{figure}[t]
  \centering
  \includegraphics[width=\textwidth]{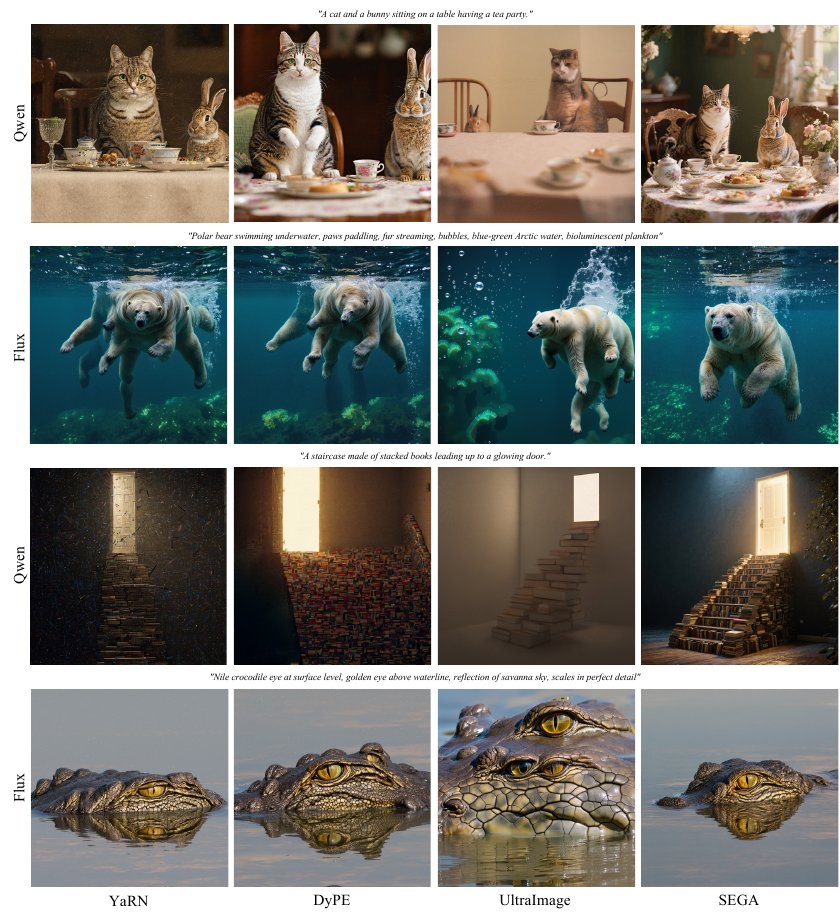}
  \vspace{-15pt}
  \caption{\textbf{Qualitative comparison (Zero-Shot Dataset).} Results on prompts from the zero-shot dataset for Qwen and Flux at $4096^2$ resolution show that SEGA handles complex environments, objects and areas with reflection, contents with challenging lighting, and preserves the shapes of the objects.
  \vspace{-10pt}}
  \label{fig:qualitative-4k-full}
\end{figure}



\begin{figure}[t]
  \centering
  \includegraphics[width=\textwidth]{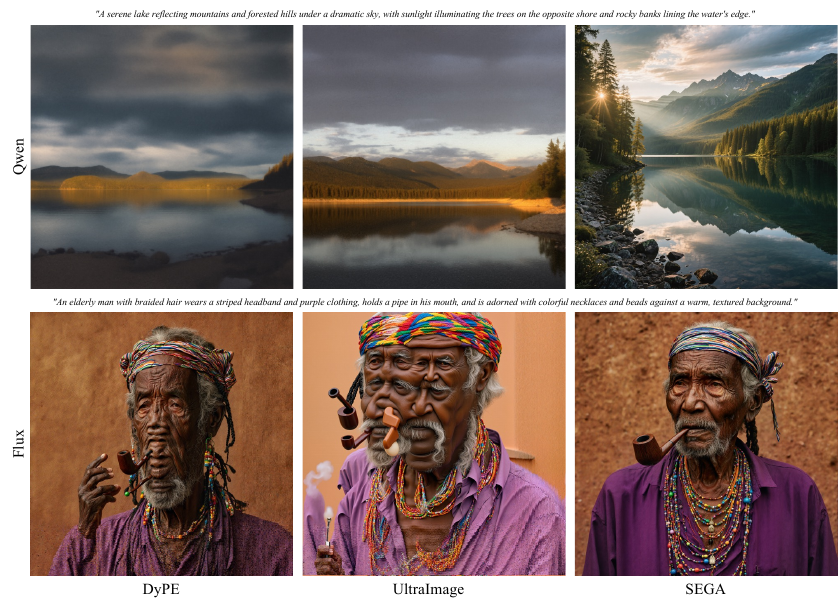}
  \vspace{-15pt}
  \caption{\textbf{Qualitative comparison (at $\mathbf{5120^2}$ resolution).} Results on two representative prompts for Qwen and Flux at $5120^2$ resolution show that SEGA elaborates on coarse and fine details as the resolution of the images increases.
  \vspace{-10pt}}
  \label{fig:qualitative-5k}
\end{figure}

\begin{figure}[t]
  \centering
  \includegraphics[width=\textwidth]{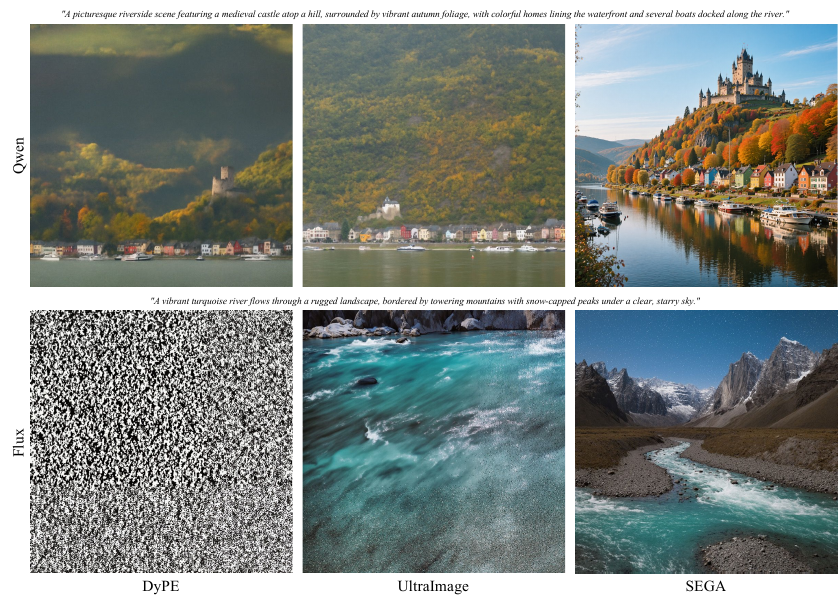}
  \vspace{-15pt}
  \caption{\textbf{Qualitative comparison (at $\mathbf{6144^2}$ resolution).} Results on two representative prompts for Qwen and Flux at $6144^2$ resolution show that SEGA makes image synthesis at this resolution possible while baselines struggle with noise and collapse of global structures.
  \vspace{-10pt}}
  \label{fig:qualitative-6k}
\end{figure}

\begin{figure}[t]
  \centering
  \includegraphics[width=\textwidth]{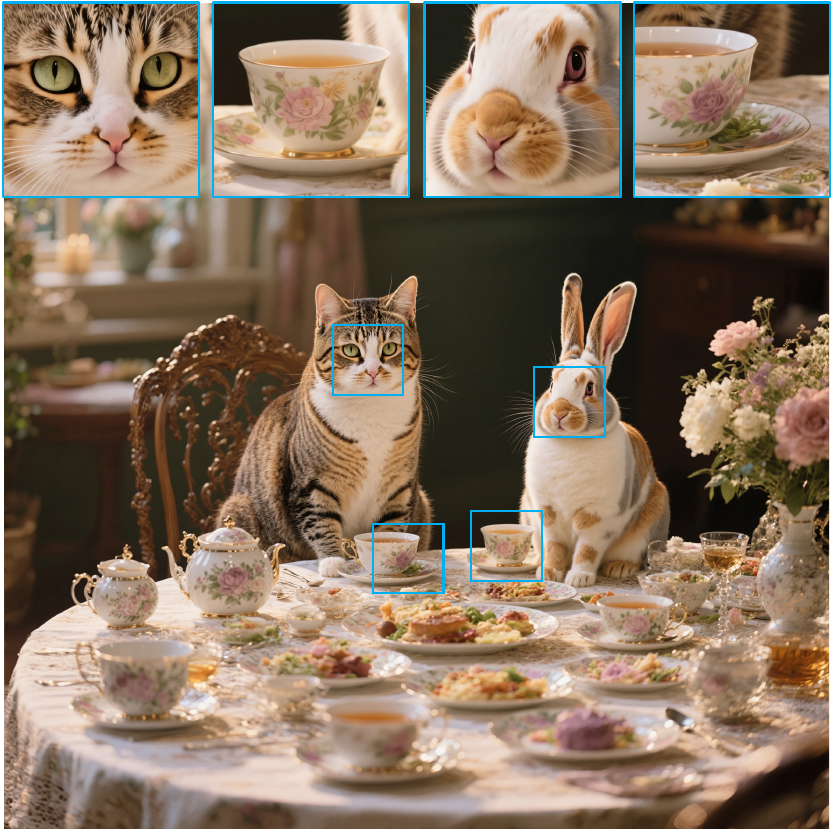}
  \vspace{-15pt}
  \caption{\textbf{Visualizing Fine-Grained Details at Extreme Resolutions.} Sample generated at $6144^2$ resolution by SEGA on Qwen. The model successfully preserves high-frequency local textures and sharp structural boundaries without experiencing structural collapse or repetition artifacts typical of long-context length extrapolation.
  \vspace{-10pt}}
  \label{fig:super_resolution}
\end{figure}

As shown in Figure~\ref{fig:qualitative-horiz-vert}, SEGA performs consistently across both vertical and horizontal aspect ratios. This figure compares YaRN, DyPE, UltraImage, and SEGA on both Flux and Qwen, demonstrating that SEGA preserves the intended image geometry without stretching or distorting objects along either spatial axis. The generated images remain sharp and visually coherent, while also maintaining strong alignment with the input prompts.

On the zero-shot prompt set, we compare YaRN, DyPE, UltraImage, and SEGA on both Flux and Qwen, shown in Figure~\ref{fig:qualitative-4k-full}. The results show that SEGA avoids common high-resolution failure modes such as repeated structures, distorted layouts, and loss of semantic clarity. In particular, SEGA maintains high prompt fidelity and fine-grained visual detail without sacrificing global composition or overall image realism.


At higher resolutions such as 5K and 6K, SEGA continues to provide clear benefits in visual sharpness, structural consistency, and prompt alignment, as shown in Figures~\ref{fig:qualitative-5k} and~\ref{fig:qualitative-6k}. These figures compare SEGA against direct-inference baselines, including DyPE and UltraImage, and demonstrate that SEGA remains effective even in challenging extrapolation regimes where other methods may produce severe artifacts or fail to generate a coherent image. For example, in Figure~\ref{fig:qualitative-6k}, DyPE fails to produce a reliable output, whereas SEGA generates a clean, consistent, and prompt-aligned image, highlighting its robustness under extreme resolution extrapolation.

Finally, Figure~\ref{fig:super_resolution} shows fine details from an ultra-high-resolution generation produced by SEGA. The zoomed-in regions illustrate that SEGA preserves local texture and object-level detail while maintaining the broader structure of the image. This suggests that SEGA's spectral-energy-guided scaling benefits both fine-scale fidelity and global coherence, rather than improving one at the expense of the other.


\section{Societal Impact and Safeguards}

Generative modeling, particularly for images and videos, has substantial potential for both beneficial and harmful use. Improvements in high-resolution generation can support creative workflows, design, visualization, and research by enabling more realistic and detailed synthesis without additional training. At the same time, increased realism may heighten risks of misuse, including disinformation, impersonation, non-consensual synthetic imagery, and amplification of existing social biases. Although SEGA does not introduce a new generative model, dataset, or training procedure, it improves the inference-time capabilities of existing text-to-image systems and may therefore amplify risks already associated with those systems. SEGA does not introduce new model-level safeguards or safety filters. Its responsible use therefore depends on the licenses, acceptable-use policies, access controls, and safety mechanisms of the underlying models and deployment platforms. In this work, we evaluate SEGA on existing models such as Flux and Qwen for research purposes. Black Forest Labs states that its Flux models and services are governed by usage policies and responsible-AI safeguards, while Qwen provides a usage policy for its AI products and services~\cite{flux,qwen}. We therefore recommend using SEGA only in ways consistent with the underlying models' licenses and usage policies, together with appropriate content moderation, provenance, and misuse-monitoring mechanisms when deployed.

%% file: tables/mix-aesthetic-main.tex
\begin{table}[t]
\centering
\caption{Comparison of SEGA against state-of-the-art baselines on SDXL~\cite{podell2023sdxl} and Diffusion-4K across four high-resolution settings on Aesthetic-4K~\cite{zhang2025diffusion}. Best and second-best results are shown in \textbf{bold} and \underline{underlined}.}
\label{tab:flux_extended}
\setlength{\tabcolsep}{4pt}
\renewcommand{\arraystretch}{1.0}
\resizebox{\textwidth}{!}{
\begin{tabular}{@{} l | cccccc | cccccc @{}}
\toprule
\multirow{2}{*}{Method} & \multicolumn{6}{c|}{2048 $\times$ 4096} & \multicolumn{6}{c}{4096 $\times$ 2048} \\
\cmidrule(lr){2-7} \cmidrule(l){8-13}
& IR$\uparrow$ & PS$\uparrow$ & CS$\uparrow$ & MSQ$\uparrow$ & FID$\downarrow$ & FID$_p$$\downarrow$
& IR$\uparrow$ & PS$\uparrow$ & CS$\uparrow$ & MSQ$\uparrow$ & FID$\downarrow$ & FID$_p$$\downarrow$ \\
\midrule
Diffusion-4K  & 0.71  & 21.89 & 27.94 & 35.76 & 156.68 & \textbf{55.95}  & 0.48  & 21.89 & 27.92 & 36.97 & \underline{154.68} & \textbf{55.03} \\
\midrule
DemoFusion    & 0.41  & 22.03 & 28.73 & 48.39 & 165.57 & 77.62  & -0.30 & 21.56 & 27.27 & \textbf{53.46} & 169.48 & 81.01 \\
FreCas        & \underline{0.93}  & 22.34 & \underline{28.89} & \textbf{55.21} & \underline{156.14} & 100.80  & 0.47  & 21.62 & 28.73 & 53.05 & 158.95 & 99.78 \\
FreeScale     & 0.89  & 22.23 & 28.78 & 50.94 & 162.86 & 98.66  & \underline{0.69}  & 21.92 & 28.97 & 44.61 & 170.32 & 94.87 \\
DiffuseHigh   & 0.73  & \underline{22.44} & 28.29 & 45.44 & 158.73 & 82.73  & 0.49  & \underline{22.18} & \underline{28.98} & 48.03 & 160.31 & 77.27 \\
\midrule
SEGA          & \textbf{1.21}  & \textbf{22.91} & \textbf{29.18} & \underline{53.65} & \textbf{151.93} & \underline{64.54}  & \textbf{0.86}  & \textbf{22.58} & \textbf{28.99} & \underline{53.30} & \textbf{153.10} & \underline{55.85} \\
\bottomrule

\toprule
\multirow{2}{*}{Method} & \multicolumn{6}{c|}{3072 $\times$ 3072} & \multicolumn{6}{c}{4096 $\times$ 4096} \\
\cmidrule(lr){2-7} \cmidrule(l){8-13}
& IR$\uparrow$ & PS$\uparrow$ & CS$\uparrow$ & MSQ$\uparrow$ & FID$\downarrow$ & FID$_p$$\downarrow$
& IR$\uparrow$ & PS$\uparrow$ & CS$\uparrow$ & MSQ$\uparrow$ & FID$\downarrow$ & FID$_p$$\downarrow$ \\
\midrule
Diffusion-4K  & 0.87  & 22.29 & 28.28 & 34.49 & 154.83 & \underline{55.79}  & 0.52  & 21.74 & 27.48 & 24.13 & 161.62 & 70.05 \\
\midrule
DemoFusion    & 0.89  & 23.00 & 29.35 & 51.45 & \underline{153.65} & 76.56  & 0.88  & 22.99 & 29.45 & \underline{41.27} & 156.94 & 72.86 \\
FreCas        & \underline{1.13}  & 23.12 & 29.20 & \underline{51.70} & 155.98 & 91.26  & \underline{1.06}  & 22.90 & \textbf{29.82} & 39.32 & \underline{156.57} & 82.27 \\
FreeScale     & 1.05  & 22.78 & \textbf{29.74} & 43.63 & 170.63 & 86.80  & \underline{1.06}  & 22.82 & \underline{29.75} & 33.94 & 167.83 & 74.72 \\
DiffuseHigh   & 0.91  & \underline{23.17} & \underline{29.59} & 49.48 & 159.33 & 73.97  & 0.93  & \underline{23.16} & 29.21 & 36.44 & 160.43 & \underline{63.19} \\
\midrule
SEGA          & \textbf{1.30}  & \textbf{23.26} & 29.29 & \textbf{52.97} & \textbf{151.08} & \textbf{43.86}  & \textbf{1.26}  & \textbf{23.18} & 29.22 & \textbf{45.72} & \textbf{150.05} & \textbf{51.28} \\
\bottomrule
\end{tabular}
}
\end{table}

%% file: tables/zero-shot.tex
\begin{table}[t]
\centering
\small
\caption{Quantitative comparison at $4096^2$ resolution on the zero-shot benchmark. Methods are grouped by backbone model; best and second-best results are \textbf{bolded} and \underline{underlined} within each group. $\dagger$ denotes a closed-source proprietary model.}
\vspace{5pt}
\label{tab:zeroshot_resolution}
\setlength{\tabcolsep}{10pt}
\begin{tabular}{ll ccccc}
\toprule
Backbone & Method & IR$\uparrow$ & PS$\uparrow$ & CS$\uparrow$ & HPS$\uparrow$ & MSQ$\uparrow$ \\
\midrule
\multirow{7}{*}{Flux}
 & Base       & -1.50 & 19.41 & 22.17 & 0.23 & 25.83         \\
 & NTK        & -0.30 & 20.66 & 25.36 & 0.25 & 29.23            \\
 & YaRN       & 0.70  & 21.83 & 28.95 & 0.27 & 41.18    \\
 & DyPE       & \underline{0.78}  & \underline{22.01} & \textbf{29.65} & 0.27 & 41.62 \\
 & UltraImage & 0.42  & 21.36 & 28.35 & \underline{0.28} & 39.29      \\
\cmidrule(l){2-7}
 & SEGA       & \textbf{1.05} & \textbf{22.50} & \underline{29.26} & \textbf{0.28} & \textbf{42.36} \\
\midrule
\multirow{4}{*}{Qwen}
 & Base       & 0.13          & 21.33 & 28.89       & 0.27          & 30.90         \\
 & DyPE       & \underline{1.13} & \underline{22.58} & 29.55 & \underline{0.29} & \underline{39.10} \\
 & UltraImage & 0.71            & 21.60 & \underline{29.62}             & 0.27            & 35.71            \\
\cmidrule(l){2-7}
 & SEGA       & \textbf{1.58}            & \textbf{23.86}  &  \textbf{30.06}           & \textbf{0.30 }           & \textbf{45.27}            \\
\midrule
Prop.$^\dagger$
 & Nano Banana 2 & 1.37 & 23.43 & 30.02 & 0.30 & 42.03 \\
\bottomrule
\end{tabular}
\end{table}

%% file: tables/aesthetic-5k.tex
\begin{table*}[t]
\centering
\scriptsize
\caption{Quantitative comparison at $5120^2$ resolution on Aesthetic-4K~\cite{zhang2025diffusion}. Methods are grouped by backbone model; best and second-best results are \textbf{bolded} and \underline{underlined} within each group.}
\vspace{5pt}
\label{tab:5k_resolution}
\resizebox{\textwidth}{!}{
\begin{tabular}{ll ccccccc}
\toprule
Backbone & Method & IR$\uparrow$ & PS$\uparrow$ & CS$\uparrow$ & HPS$\uparrow$ & MSQ$\uparrow$ & CQA$\uparrow$ & FID$\downarrow$ \\
\midrule
\multirow{6}{*}{Flux}
 & Base       & -2.03 & 18.22 & 16.43 & 0.21 & 25.72    & 0.37            & 351.73     \\
 & NTK        & -0.05 & 21.31 & 24.53 & 0.25 & 27.96    & 0.48            & 244.52     \\
 & YaRN       &  0.40 & 21.53 & 26.74 & 0.26 & 40.11    & 0.61            & 235.33     \\
 & DyPE       & \underline{0.58} & \underline{21.85} & \underline{27.98} & 0.26 & \underline{40.40} & 0.66 & \underline{225.30} \\
 & UltraImage &  0.24 & 21.34 & 26.39 & \underline{0.27} & 36.25 & \textbf{0.74} & 251.02 \\
\cmidrule(l){2-9}
 & SEGA       & \textbf{1.13} & \textbf{23.22} & \textbf{28.92} & \textbf{0.29} & \textbf{40.64} & \underline{0.72} & \textbf{221.99} \\
\midrule
\multirow{4}{*}{Qwen}
 & Base       & -0.54 & 20.08 & 25.52 & 0.24 & 21.90    & 0.48            & 270.94     \\
 & DyPE       & -0.37 & 20.34 & 24.72 & 0.24 & 20.58 & 0.41          & \underline{250.45} \\
 & UltraImage & \underline{0.18} & \underline{20.88} & \underline{26.13} & \underline{0.25} & \underline{23.13} & \underline{0.49} & 255.24 \\
\cmidrule(l){2-9}
 & SEGA       & \textbf{1.39} & \textbf{23.94} & \textbf{29.55} & \textbf{0.30} & \textbf{41.62} & \textbf{0.74} & \textbf{218.47} \\
\bottomrule
\end{tabular}
}
\end{table*}

%% file: tables/aesthetic-6k.tex
\begin{table*}[t]
\centering
\scriptsize
\caption{Quantitative comparison at $6144^2$ resolution on Aesthetic-4K~\cite{zhang2025diffusion}. Methods are grouped by backbone model; best and second-best results are \textbf{bolded} and \underline{underlined} within each group.}
\vspace{5pt}
\label{tab:6k_resolution}
\resizebox{\textwidth}{!}{
\begin{tabular}{ll ccccccc}
\toprule
Backbone & Method & IR$\uparrow$ & PS$\uparrow$ & CS$\uparrow$ & HPS$\uparrow$ & MSQ$\uparrow$ & CQA$\uparrow$ & FID$\downarrow$ \\
\midrule
\multirow{6}{*}{Flux}
 & Base       & -2.25 & 17.06 & 12.48 & 0.19 & 25.77    & 0.31            & 453.97     \\
 & NTK        & -2.03 & 17.11 & 10.83 & 0.19 & 23.79    & 0.34            & 528.45     \\
 & YaRN       & \underline{-0.33} & 20.18 & 24.16 & 0.24 & 35.28    & 0.56   & 288.66     \\
 & DyPE       & -2.23 & 16.78 & 11.68 & 0.18 & 27.65 & 0.34          & \underline{274.82} \\
 & UltraImage & -0.74 & \underline{20.27} & \underline{25.29} & \underline{0.26} & \underline{36.15} & \textbf{0.69} & 290.75 \\
\cmidrule(l){2-9}
 & SEGA       & \textbf{0.75} & \textbf{22.47} & \textbf{27.92} & \textbf{0.27} & \textbf{43.43} & \underline{0.65} & \textbf{232.18} \\
\midrule
\multirow{4}{*}{Qwen}
 & Base       & -0.95 & 19.78 & \underline{23.96} & 0.23 & 22.50    & \underline{0.44}           & 279.78     \\
 & DyPE       & -0.88 & 19.64 & 23.75 & 0.23 & \underline{36.15} & 0.35 & 279.77 \\
 & UltraImage & \underline{-0.47} & \underline{20.14} & 23.94 & \underline{0.24} & 21.89 & 0.43 & \underline{254.09} \\
\cmidrule(l){2-9}
 & SEGA       & \textbf{1.36} & \textbf{23.97} & \textbf{28.75} & \textbf{0.30} & \textbf{38.77} & \textbf{0.74} & \textbf{210.21} \\
\bottomrule
\end{tabular}
}
\end{table*}